\pdfoutput=1

\documentclass[11pt]{article}

\usepackage[]{EMNLP2022}

\usepackage{times}
\usepackage{latexsym}

\usepackage[T1]{fontenc}

\usepackage[utf8]{inputenc}

\usepackage{microtype}
\usepackage{graphics}
\usepackage{graphicx}

\usepackage{algorithm}
\usepackage{algorithmic}

\usepackage{amsmath,amsfonts,bm}
\usepackage{xspace}
\usepackage{mathptmx}
\usepackage{bbm}

\usepackage{CJKutf8}
\usepackage{booktabs}
\usepackage{multirow}
\usepackage{verbatim}
\usepackage{subcaption}
\usepackage{caption}
\title{TemporalWiki: A Lifelong Benchmark for Training and Evaluating Ever-Evolving Language Models}

\author{Joel Jang{\textsuperscript{1,}\thanks{\;\;indicates equal contribution.}} \quad Seonghyeon Ye{\textsuperscript{1,$\ast$}} \quad Changho Lee{\textsuperscript{3}}\quad Sohee Yang{\textsuperscript{1}} \\ {\bf Joongbo Shin{\textsuperscript{2}} \quad  Janghoon Han{\textsuperscript{2}} \quad Gyeonghun Kim{\textsuperscript{2}} \quad Minjoon Seo{\textsuperscript{1}}} \\
        {\textsuperscript{1}}KAIST\quad{\textsuperscript{2}}LG AI Research\quad{\textsuperscript{3}}Korea University \\ 
        \texttt{\{joeljang,seonghyeon.ye,sohee.yang,minjoon\}@kaist.ac.kr}\\ \texttt{ckdgh0801@korea.ac.kr} \quad
        \texttt{\{jb.shin,janghoon.han,ghkayne.kim\}@lgresearch.ai}
        }

\begin{document}
\maketitle
\begin{abstract}
Language Models (LMs) become outdated as the world changes; they often fail to perform tasks requiring recent factual information which was absent or different during training, a phenomenon called \textit{temporal misalignment}. This is especially a challenging problem because the research community still lacks a coherent dataset for assessing the adaptability of LMs to frequently-updated knowledge corpus such as Wikipedia. To this end, we introduce \textsc{TemporalWiki}, a lifelong benchmark for ever-evolving LMs that utilizes the difference between consecutive snapshots of English Wikipedia and English Wikidata for training and evaluation, respectively. The benchmark hence allows researchers to periodically track an LM's ability to retain previous knowledge and acquire updated/new knowledge at each point in time. We also find that training an LM on the \emph{diff} data through continual learning methods achieves similar or better perplexity than on the entire snapshot in our benchmark with 12 times less computational cost, which verifies that factual knowledge in LMs can be safely updated with minimal training data via continual learning. The dataset and the code is made available at \href{https://github.com/joeljang/temporalwiki}{this link}. 
\end{abstract}

\section{Introduction}
\label{sec:introduction}
\begin{figure*}[t!]
    \centering
    \includegraphics[width=0.75\textwidth]{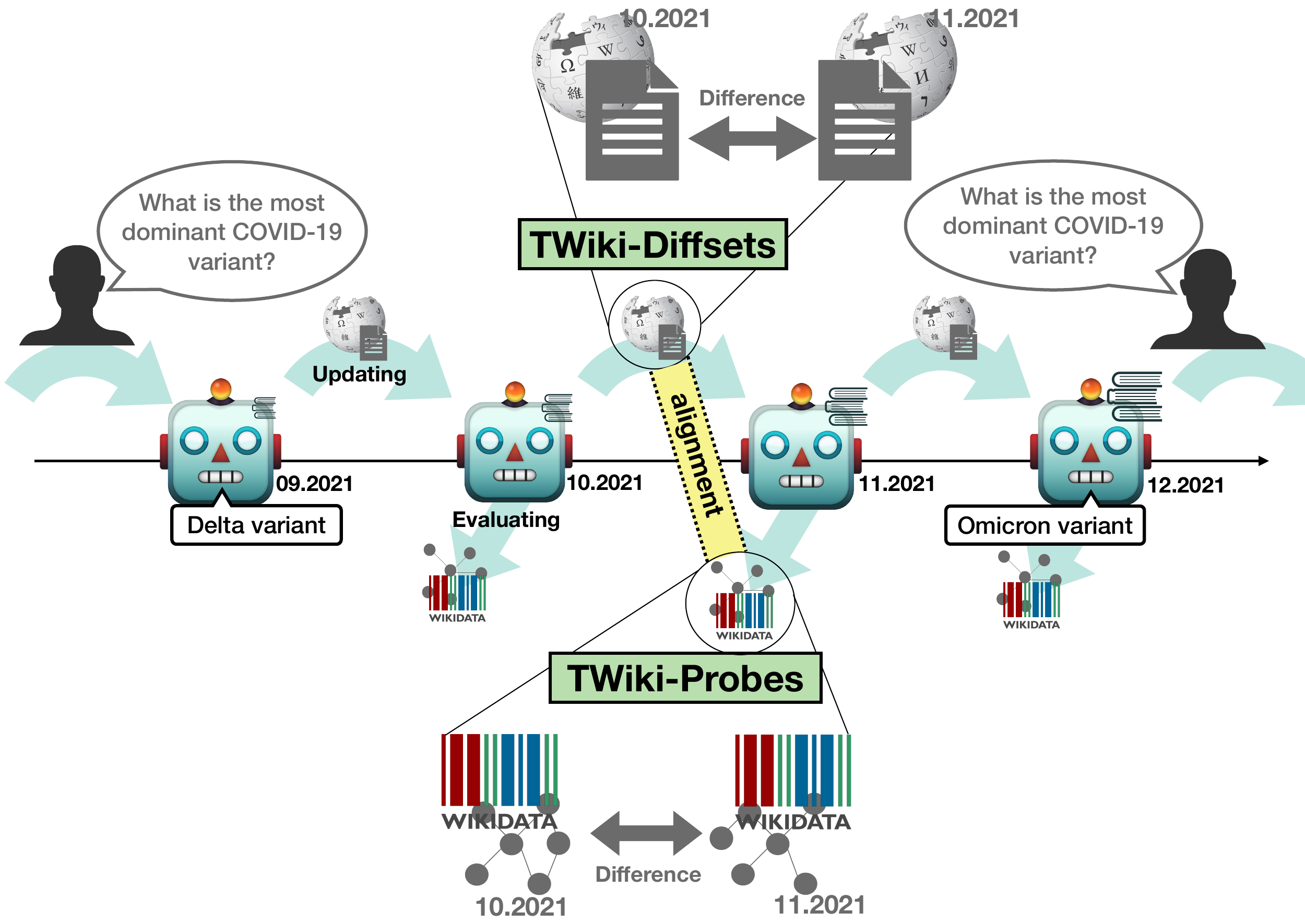}
    \caption{An overview of using \textsc{TemporalWiki}, consisting of \textsc{TWiki-Diffsets} and \textsc{TWiki-Probes} to train and evaluate ever-evolving LMs, respectively. Differences between Wikipedia snapshots at different points in time are used for temporal language modeling, and categorized factual instances in the corresponding Wikidata snapshots are used for temporal evaluation.}
    \label{fig:fig1}
\vspace{-5mm}
\end{figure*}

Large Language Models (LMs) pretrained on a vast amount of text corpus have shown to be highly effective when finetuned or prompted to perform various downstream tasks~\citep{raffel2019exploring, brown2020language, sanh2021multitask, wei2021finetuned}. However, most of the datasets used to evaluate these LMs are static benchmarks; the train and test data are both from similar points in time. On the other hand, in the real world, factual knowledge is frequently changed, added, or deprecated. For example, suppose a language model is asked  what the most dominant coronavirus variant is (Figure~\ref{fig:fig1}).  The answer would have been the \textit{Delta variant} in the fall of 2021 but has changed to the \textit{Omicron variant} near the end of 2021. If LMs remain unchanged and are not periodically trained to cope with the changing world, they will be outdated very quickly. This means downstream tasks that directly depend on or are finetuned from the LM will suffer from \textit{temporal misalignment}~\citep{luu2021time, lazaridou2021mind}, which refers to the misalignment in time between the train and test data.

Temporal misalignment becomes a critical problem, especially when using language models for knowledge-intensive tasks such as closed-book question answering~\citep{roberts2020much, petroni2020kilt, jang2021towards} since they rely solely on the knowledge stored in their parameters. Furthermore, LMs augmented with retrieval mechanism~\citep{guu2020realm, lewis2020retrieval, borgeaud2021improving} often suffer from \textit{hallucination} even if they successfully retrieve up-to-date information~\citep{zhang2021situatedqa, chen2021dataset,longpre-etal-2021-entity}. This means that the implicit knowledge stored in the model parameters has to be updated as well because it may cause conflicts with the explicit knowledge retrieved from external sources such as up-to-date knowledge bases and ultimately cause the LM to \textit{hallucinate}.

Recently, \citet{lazaridou2021mind, jang2021towards} have explored updating the internal knowledge of LMs through continual pretraining on new and updated data as a solution for mitigating temporal misalignment. However, these datasets are still \textit{static} in nature: as the world changes, they will eventually get outdated as well. In order to comprehensively measure the capability of ever-evolving LMs on addressing temporal misalignment, automated periodic evaluation of the LMs is crucial.

In this paper, we introduce \textsc{TemporalWiki}, a \textit{lifelong} benchmark for training and evaluating ever-evolving LMs in a periodic and automated manner, shown in Figure \ref{fig:fig1}. The corpora used for updating LMs are constructed by comparing articles from consecutive English Wikipedia snapshots and retrieving only \textit{changed} information, which  we name as \textsc{TWiki-Diffsets}. The evaluation datasets are constructed in a similar manner by comparing English Wikidata snapshots that correspond to the Wikipedia snapshots in time and categorizing each factual instance into \textsc{Unchanged} or \textsc{Changed}. Since Wikidata updates may not exactly align with Wikipedia updates, we only retain factual instances that can be grounded to articles in Wikipedia, ensuring the quality of the data and name the resulting evaluation dataset as \textsc{TWiki-Probes}. The entire benchmark creation process is done without any human annotation, thus allowing it to be automated and \textit{lifelong} as new English Wikipedia and English Wikidata snapshots are released by Wikimedia\footnote{\href{https://commons.wikimedia.org/}{https://commons.wikimedia.org/}} on a monthly basis.

Through \textsc{TemporalWiki}, we aim to tackle the following research questions: How can we train ever-evolving LMs efficiently and automate the evaluation of each update? How does updating LMs only on updated data from Wikipedia compare to updating LMs on entire Wikipedia snapshots, especially in scenarios requiring multiple updates? How problematic is catastrophic forgetting~\citep{mccloskey1989catastrophic} when LMs are updated only on \textit{new} data, and how can we effectively mitigate catastrophic forgetting? Our main contributions are summarized as follows: 

\begin{itemize}
  \item We introduce \textsc{TemporalWiki}, a \textit{lifelong} benchmark for ever-evolving LMs. Unlike previous \textit{static} benchmarks, \textsc{TemporalWiki} is responsive to the \textit{dynamic} changes in the world and can be utilized to automatically train and evaluate ever-evolving LMs on each English Wikipedia and English Wikidata snapshot update. 
  \item We find that continually training LMs only on the updated portion of English Wikipedia, which we call \emph{temporal language modeling}, is much more efficient than updating LMs on entire English Wikipedia snapshots in terms of both computation and stability-plasticity trade-off. It is still a challenging task, especially when multiple updates are required due to catastrophic forgetting. 
  \item As competitive baselines for temporal language modeling, we implement previous continual learning approaches that mitigate forgetting while bolstering the learning of new knowledge, thus providing an overall enhancement in terms of both stability and plasticity. 
\end{itemize}
We hope that \textsc{TemporalWiki} will foster future research towards training ever-evolving LMs.

\section{Background}
Recent works have introduced the need to tackle the issue of temporal misalignment, which refers to neural networks showing poor performance due to misalignment in time between the train and test data. Temporal misalignment can be caused either by (1) the dynamic nature of language~\citep{rottger2021temporal,hombaiah2021dynamic, rosin2021time, loureiro2022timelms} or (2) the update of factual information~\citep{chen2021dataset, dhingra2021time, jang2021towards}.

\citet{luu2021time} have emphasized the effect of temporal misalignment on eight different NLP downstream tasks, asserting that misalignment between the train and test sets of the downstream tasks causes severe performance degradation that can be mitigate finetuning on the corpus from the target period. \citet{agarwal2021temporal} have argued this to be less of a concern when utilizing representations from pretrained LMs and show that self-labeling on the downstream task is more effective than continued pretraining on more recent data for temporal adaptation. Note that these works have focused on misalignment caused by the dynamic nature of language on tasks that are not knowledge-intensive, such as text classification. 

Others have tackled the problem caused by the update of factual knowledge. \citet{lazaridou2021mind} have shown that LMs deteriorate significantly in performance when there is a misalignment in time between the pretraining data and the downstream task and argued ever-evolving LMs are necessary. \citet{dhingra2021time} have proposed explicitly including time information during pretraining as a potential solution. \citet{jang2021towards, jin2021lifelong} have implemented continual learning methods to mitigate catastrophic forgetting that occurs during continued pretraining on new data. 

Despite the recent community interest in the need for ever-evolving LMs, the community lacks widely-available resources to train and evaluate such LMs. Previous works have introduced benchmarks comprised of data sources from Twitter feeds~\citep{osborne-etal-2014-exponential, yogatama2014dynamic, loureiro2022timelms}, recent news articles~\citep{jang2021towards}, and arXiv papers~\citep{lazaridou2021mind} where the temporal adaptability of LMs and the effectiveness of different methodologies of updating LMs can be evaluated. However, these data sources are domain-specific and inherently \textit{static}. 

On the other hand, Wikipedia and Wikidata are known to be great sources of general world knowledge and thus have been widely used by the community~\citep{dinan2018wizard, thorne2018fever, kwiatkowski2019natural, piktus2021web}. 120K volunteer editors make 120 updates to the English Wikipedia per minute and add hundreds of new article entries every day~\citep{logan2021fruit}\footnote{\href{https://en.wikipedia.org/wiki/Wikipedia:Statistics}{https://en.wikipedia.org/wiki/Wikipedia:Statistics}}. Even though every Wikipedia and Wikidata update may not correspond to an actual change in the \textit{real} world, \textsc{TemporalWiki} leverages the dynamic nature of Wikipedia and Wikidata to provide a \emph{lifelong} benchmark for developing and maintaining ever-evolving LMs.

\section{TemporalWiki}
In this section, we delve into the process of creating \textsc{TemporalWiki}, which is comprised of training corpora (\textsc{TWiki-Diffsets}) and evaluation datasets (\textsc{TWiki-Probes}) sourced from English Wikipedia and English Wikidata, respectively. For efficiency, \textit{English} is abbreviated when referring to English Wikipedia and English Wikidata throughout the paper. Moreover, we clarify that not all Wikipedia/Wikidata updates equate to actual updates of \textit{world} knowledge. In Section \ref{sec:train}, we first describe the process of constructing the training corpora from Wikipedia snapshots. Then in Section \ref{sec:eval}, we describe the process of generating the evaluation datasets from Wikidata snapshots. In Section \ref{sec:qual}, we describe the quality control applied to the evaluation datasets.

\begin{algorithm}[h!]
\small
\caption{\small Generating \textsc{TWiki-Diffsets}}
\begin{algorithmic}
    \REQUIRE Wikipedia snapshots $WP_{prev}$ and $WP_{recent}$ where $WP_{recent}$ is more recent.
    \STATE $D$ := An empty array to store new and updated data.
    \STATE *$article$ in $WP$ has attributes $id$ and $text$
    \FORALL{$article$ $a_{r} \in WP_{recent}$}
        \IF{$a_{r}.id = a_{p}.id$ for some $article$ $a_{p} \in WP_{prev}$}
            \STATE $D$.append($\textsc{GetDiff}(a_{p}, a_{r}$))
        \ELSE
            \STATE $D$.append($a_{r}$)
        \ENDIF
    \ENDFOR
\STATE
\STATE \textbf{function} \textsc{GetDiff}($a_{p}, a_{r}$)
    \STATE $Diff$ := An empty string to append difference between $text$ in two $article$s.
    \FORALL{paragraph $p_{r} \in a_{r}.text$}
        \IF{$p_{r}$ have \textit{no} matching sentences with any paragraph $p_{p} \in a_{p}.text$ }
            \STATE $Diff$ $\leftarrow Diff + p_{r}$
        \ELSIF{$p_{r}$ have \textit{some} matching and \textit{some} different sentences with any paragraph $p_{p} \in a_{p}.text$ }
            \STATE $Diff$ $\leftarrow Diff$ + \textit{sentences} that differ between $p_{r}$ and $p_{p}$.
        \ENDIF
    \ENDFOR
    \STATE \textbf{return} $Diff$
\end{algorithmic}
\label{algo1}
\end{algorithm}

\subsection{Generating Training Corpora from Wikipedia}
\label{sec:train}

It is highly computationally expensive to train an LM on the entire Wikipedia snapshot every time the LM requires updates since most part of Wikipedia is \textit{unchanged} from the previous snapshot. Moreover, it is not certain whether training on whole snapshot is the best approach for updating the factual knowledge stored in the LM. Therefore, we compare the differences between consecutive Wikipedia snapshots in order to use only updated and new text for training. We call these subsets \textsc{TWiki-Diffsets}. Algorithm \ref{algo1} shows the procedure for generating them. 

As shown in Algorithm \ref{algo1}, a single \textsc{TWiki-Diffset} is generated by getting the differences (similarly to \texttt{git diff}) between two consecutive Wikipedia snapshots. If an article with a new unique id is included in the recent snapshot, we append the entire article to \textsc{TWiki-Diffset}. For an article having an existing id in the previous snapshot, we compare the two articles by paragraphs and add new or updated sentences to \textsc{TWiki-Diffset}. Examples of \textsc{TWiki-Diffset} are shown in Figure \ref{fig:twiki_diffsets}, and detailed statistics are shown in Section \ref{sec:statistics}.


\subsection{Generating Evaluation Datasets from Wikidata}
\label{sec:eval}
The success of a LM update for continual pretraining setting can be evaluated by quantifying the stability-plasticity dilemma~\citep{mermillod2013stability}: the dilemma of neural models having to sacrifice either \textit{stability}, ability to retain learned knowledge, or \textit{plasticity}, ability to obtain new knowledge. In order to evaluate whether each update is successful, we need evaluation datasets that can quantify the amount of \textit{changed} (updated or new) knowledge successfully gained (plasticity) and the amount of knowledge that remains \textit{unchanged} as intended (stability). Therefore, we categorize factual instances from Wikidata snapshots that are temporally aligned with Wikipedia snapshots and call the resulting datasets \textsc{TWiki-Probes}.

Wikidata snapshots are structured knowledge graphs that store factual information in the form of (\texttt{Subject}, \texttt{Relation}, \texttt{Object}) such as (\texttt{Barack Obama}, \texttt{born-in}, \texttt{Hawaii}). These factual instances can be used to probe the LM for factual knowledge~\citep{petroni2019language}. Through Algorithm \ref{algo2}, we distinguish each factual instance into either \textsc{Unchanged} or \textsc{Changed}.

\begin{algorithm}
\caption{\small Generating \textsc{TWiki-Probes}}
\begin{algorithmic} 
\small
    \REQUIRE Wikidata snapshots $WD_{prev}$ and $WD_{recent}$ where $WD_{recent}$ is more recent. 
    \STATE $Un$, $C$ := Arrays that store \textsc{Unchanged} and \textsc{Changed} factual instances, respectively.
    \FORALL{fact $(s_r, r_r, o_r) \in WD_{recent}$}
        \STATE $\mathbb{P} \leftarrow \{(s, r, o) \mid s=s_r \text{ where}\ (s,r,o) \in WD_{prev} \}$
        \IF{$\mathbb{P} = \emptyset$}
            \STATE $C$.append($s_r, r_r, o_r$)
        \ELSIF{$r_r \notin \mathbb{P}$}
            \STATE $C$.append($s_r, r_r, o_r$)
        \ELSIF{$r = r_r \text{ and } o = o_r \text{ for some} (s,r,o) \in \mathbb{P}$}
            \STATE $Un$.append($s_r, r_r, o_r$)
        \ELSE
            \STATE $C$.append($s_r, r_r, o_r$)
        \ENDIF
    \ENDFOR
\end{algorithmic}
\label{algo2}
\end{algorithm}

As shown in Algorithm \ref{algo2}, given two consecutive Wikidata snapshots, a single \textsc{TWiki-Probe} is constructed, which is used to evaluate an LM updated with \textsc{TWiki-Diffset}. Algorithm \ref{algo2} categorizes instances with new \texttt{Relation} or instances with the same \texttt{Relation}, but a new \texttt{Object} into \textsc{Changed}, and unchanged instances into \textsc{Unchanged}.

\subsection{Quality Control for Evaluation Data}
\label{sec:qual}
We apply several quality control steps to the categorized factual instances from Section~\ref{sec:eval} to reflect the actual knowledge change from the LM update.

\paragraph{Alignment with \textsc{TWiki-Diffsets}} We ensure correct alignment of \textsc{Changed} instances with articles in \textsc{TWiki-Diffsets} and \textsc{Unchanged} instances with articles from the entire Wikipedia since Wikidata updates do not necessarily entail Wikipedia updates and vice versa. In order to do this, we take three steps. \textbf{Step \#1}: We crawl information from each Wikipedia article page to find the mapping to the corresponding Wikidata entity id and store the information as a dictionary. \textbf{Step \#2}: Then, for each factual instance from \textsc{Changed}, we check if the \texttt{Subject} id can be mapped to an article from \textsc{TWiki-Diffsets} using the dictionary of id mappings. Likewise, for each instance from \textsc{Unchanged}, we check if the \texttt{Subject} id can be mapped to an article from Wikipedia. \textbf{Step \#3}: Lastly, for a successfully mapped factual instance from Step 2, we finally keep the instances where \texttt{Object} exists in the text of the article. 

\paragraph{Heuristic Filtering} In addition to the alignment with \textsc{TWiki-Diffsets}, in order to further ensure the quality of the evaluation datasets, we apply three heuristic filtering rules to strengthen the quality of the data. \textbf{Rule \#1}: We remove the instances where either \textsc{Subject} or \textsc{Object} is a substring of the other. \textbf{Rule \#2}: We remove the instances where \textsc{Object} contains more than 5 words. \textbf{Rule \#3}: We limit the proportion of single \textsc{Subject} to have 1\% of the total, and \textsc{Relation} and \textsc{Object} by 5\% of the total. Table \ref{table:twiki_probes_example} shows some examples of \textsc{TWiki-Probes} after quality control.



\section{Dataset Statistics}
\label{sec:statistics}
In this paper, we construct \textsc{TemporalWiki} from 08.2021 to 12.2021\footnote{As new Wikipedia and Wikidata dumps are available on a monthly basis, we provide the source code for constructing new \textsc{TWiki-Diffsets} and \textsc{TWiki-Probes} at \href{https://github.com/joeljang/temporalwiki}{this link}} and its statistics are discussed below.

\paragraph{Training Corpora Statistics}  Statistics of Wikipedia snapshots and \textsc{TWiki-Diffsets} are shown in Table \ref{table:twiki_stats}. An interesting aspect of \textsc{TWiki-Diffsets} is that the amount of information being updated and added (i.e., number of tokens in each subset) is similar for each month. 

\begin{table}[t!]
    \centering
    \fontsize{10}{12}\selectfont
        \caption{Statistics of \textsc{TWiki-Diffsets}. The two digits indicate the month of the year 2021 that the Wikipedia snapshot was obtained from. The four digits for \textsc{Wiki-Diffset} indicate the months of the two snapshots being compared. For instance, \textsc{TWiki-Diffset-0809} indicates the difference between August (08) and September (09).}
    \resizebox{0.85\columnwidth}{!}{\begin{tabular}{lrr}
    \toprule
                      & \textbf{\# of Articles} & \textbf{\# of Tokens} \\
    \midrule
    \textsc{Wikipedia-08}     & 6.3M    & 4.6B       \\
    \textsc{TWiki-Diffset-0809} & 306.4K     & 347.29M      \\
    \textsc{Wikipedia-09}     & 6.3M      & 4.6B       \\
    \textsc{TWiki-Diffset-0910} & 299.2K & 347.96M         \\
    \textsc{Wikipedia-10}     & 6.3M      & 4.7B       \\
    \textsc{TWiki-Diffset-1011} & 301.1K    & 346.45M         \\
    \textsc{Wikipedia-11}     & 6.3M      & 4.6B       \\
    \textsc{TWiki-Diffset-1112} & 328.9K & 376.09M      \\
    \textsc{Wikipedia-12}     & 6.3M  & 4.7B     \\
    \bottomrule
    \end{tabular}}
    \vspace{-2mm}
    \label{table:twiki_stats}
\end{table}

\begin{table}[t!]
    \caption{Detailed Statistics of \textsc{TWiki-Probes} during construction. \textbf{Un} and \textbf{C} represents \textsc{Unchanged} and \textsc{Changed} factual instances, respectively.}
    \resizebox{\columnwidth}{!}{\begin{tabular}{ccccccccc}
    \toprule
        \multicolumn{1}{l}{} & \multicolumn{2}{c}{\textbf{Initial Categorization}} & \multirow{2}{*}{$\rightarrow$} & \multicolumn{2}{c}{\textbf{Alignment}}& 
        \multirow{2}{*}{$\rightarrow$} &
        \multicolumn{2}{c}{\textbf{Heuristic Filtering}} \\
        \cmidrule(lr){2-3} \cmidrule(lr){5-6} \cmidrule(lr){8-9}
        \textbf{Month}
        & \textbf{Un} & \textbf{C} &       
        & \textbf{Un} & \textbf{C} &
        & \textbf{Un} & \textbf{C} \\
    \midrule
    \textsc{0809} & 514,017 & 1,209,272 & & 10,133 & 2,329 & & 6,935 & 1,776 \\
    \textsc{0910} & 544,708 & 1,196,806 & & 10,625 & 2,621 & & 7,340 & 1,982 \\
    \textsc{1011} & 460,228 & 1,572,778 & & 10,544 & 1,742 & & 7,313 & 1,358 \\
    \textsc{1112} & 463,623 & 1,653,709 & & 10,580 & 3,472 & & 7,293 & 1,951 \\
    \bottomrule
    \end{tabular}}
\label{table:probes_stats}
\vspace{-5mm}
\end{table}

\paragraph{Evaluation Dataset Statistics} The statistics of \textsc{TWiki-Probes} from the initial categorization from Algorithm \ref{algo2} and quality control are shown in Table \ref{table:probes_stats}\footnote{A single Wikidata snapshot is comprised of 93 million distinct entities, where there are around 30 facts for each entity which amounts to roughly \textit{2.8 billion factual instances}. Since most instances from Algorithm \ref{algo2} are categorized into \textsc{Unchanged}, we randomly sample 0.1\% of the factual instances after applying Algorithm \ref{algo2}.}. For further analysis, we break down the entity types of \texttt{Subject} and \texttt{Object}, and observe a similar proportion of each entity category for each month of \textsc{TWiki-Probes} (Appendix \ref{appen:entity_type}). We also show the distribution of the top 30 most frequent \texttt{Relation} of \textsc{Unchanged} and \textsc{Changed} (Appendix \ref{appen:relation}). 

\section{Experiments with \textsc{TemporalWiki}}
In this section, we train and evaluate ever-evolving LMs with \textsc{TemporalWiki}. Section \ref{sec:settings} describes the experimental settings. Section \ref{sec:baselines} describes the baseline methodologies for updating LMs. Section \ref{sec:intrinsic} shows evaluation results on the training corpora. Section \ref{sec:extrinsic} presents the experimental results on \textsc{TWiki-Probes}.

\subsection{Experimental Settings}
\label{sec:settings}
For our baseline language model (LM), we continue pretraining GPT-2 Large~\citep{radford2019language} (774M parameters).
We first compare the baseline performances between updating GPT-2 with \textsc{TWiki-Diffsets} and updating it with entire Wikipedia snapshots and evaluate each update using \textsc{TWiki-Probes}. We also implement continual learning methods from literature known for mitigating \textit{catastrophic forgetting} that occurs when updating GPT-2 with only \textsc{TWiki-Diffsets}. Further detailed configuration of the experimental settings is provided in Appendix \ref{appen:config}.

\subsection{Baseline Models}
\label{sec:baselines}
Here we describe the baseline methods used for training and evaluation, namely \textsc{Initial}, \textsc{Full}, \textsc{Diff}, \textsc{RecAdam}, \textsc{Mix-review}, \textsc{K-Adapter}, and \textsc{LoRA}.

\paragraph{Initial} As the starting model checkpoint for all of the experiments, we continually pretrain pretrained GPT-2 from \citet{radford2019language} on the 08.2021 Wikipedia snapshot for four epochs in total (around 546K global steps) so that the initial GPT-2 used for all of the experiments is updated with the last two years of world knowledge. We denote this checkpoint as \textsc{Initial}, and it serves as the initial checkpoint for all of the other methods. 

\paragraph{Full} We start from \textsc{Initial} and continue pretraining it on the entire Wikipedia snapshot of each month in a sequential manner. For example, after training on the 09.2021 Wikipedia snapshot from \textsc{Initial}, we continue training it on the 10.2021  Wikipedia snapshot and move on to the next snapshot. We denote the resulting model as \textsc{Full}. We iterate through the training data only once, which corresponds to an average of 4.6 billion token updates (140K global steps) for each month.

\paragraph{Diff} We start from \textsc{Initial} and continue pretraining it on \textsc{TWiki-Diffsets} in a sequential manner. We denote the resulting model as \textsc{Diff}. Similarly to \textsc{Full}, we iterate through the training data only once, which is an average of 347 million token updates (12K global steps) for each month.

\paragraph{RecAdam} We implement a \textit{regularization-based} continual learning method for training large LMs called \textsc{RecAdam}~\citep{chen2020recall} which places a stronger independent assumption among the model parameters, overcoming the limitations of implementing traditional methods such as EWC~\citep{kirkpatrick2017overcoming} for training large language models. We set the hyperparameters of the optimizer identical to the original implementation. 

\paragraph{Mix-review} We implement a \textit{rehearsal-based} continual learning method for training large LMs called \textsc{Mix-review}~\citep{he2021analyzing} which mixes in random subsets of the initial pretraining data (08.2021 Wikipedia data). We fix the mix-ratio as 2 in our experiments. 

\paragraph{LoRA} We implement a parameter-expansion-based continual learning method called \textsc{LoRA}~\citep{hu2021lora} which freezes the original parameters while adding trainable rank-decomposition matrices into each layer. We use hyperparameters identical to the optimal setting of the original implementation.

\paragraph{K-Adapter}
We implement another parameter-expansion-based continual learning method, \textsc{K-Adapter}~\citep{wang2020k}, which freezes the original parameters while adding additional adapters (an increase of 103M parameters) to the LM. \footnote{We add the additional parameters once for the updates from 08.2021. Exploring the optimal interval to add parameters for ever-evolving LMs is left for future work.}

\subsection{Intrinsic Evaluation}
\label{sec:intrinsic}
\begin{figure}[t!]
    \centering
    \begin{subfigure}[b]{0.4\textwidth}
    \includegraphics[width=\textwidth]{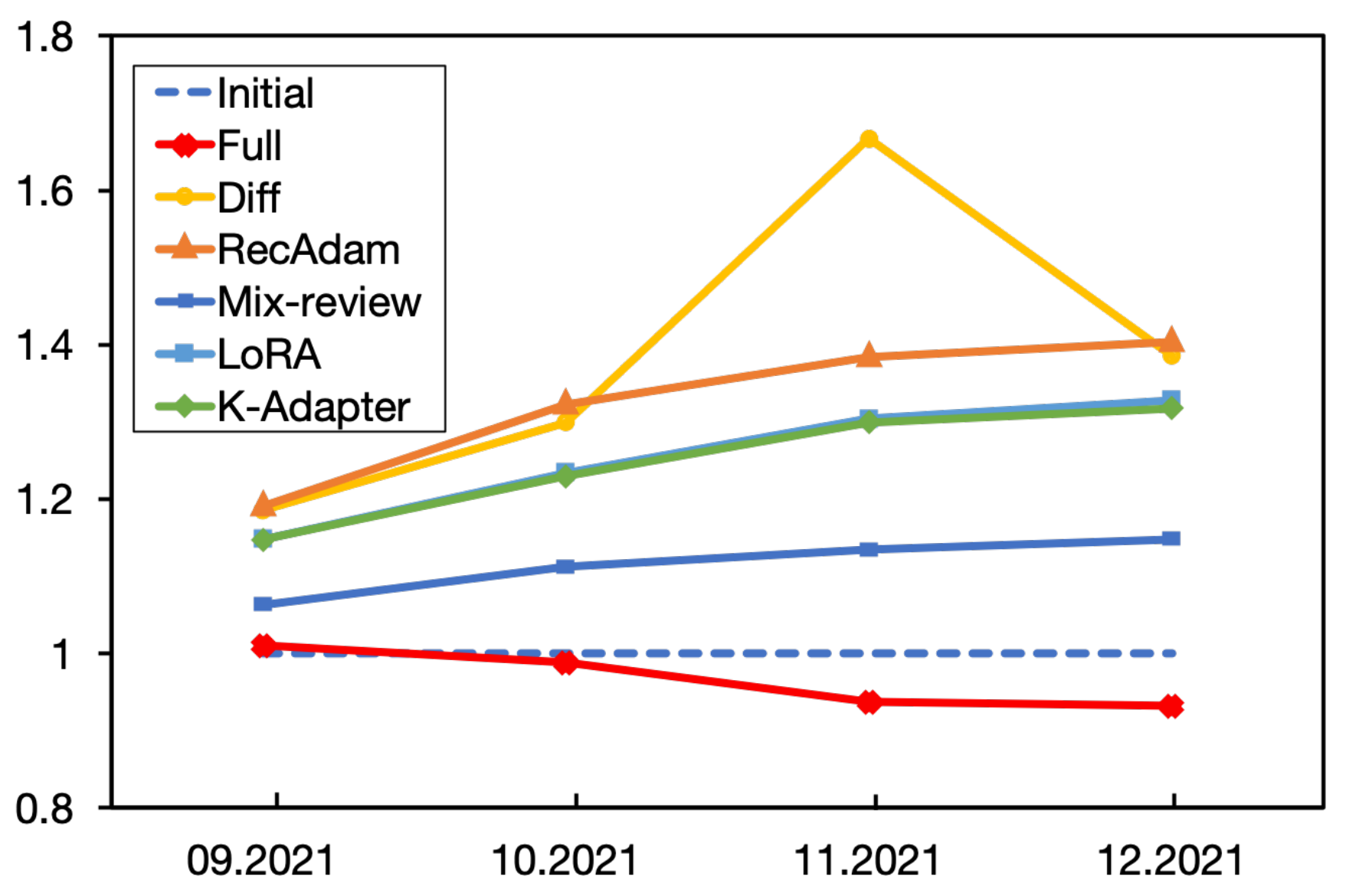} 
    \caption{\textsc{Non-TWiki-Diffsets}}
    \end{subfigure}
    \hspace{2em}
    \begin{subfigure}[b]{0.4\textwidth}
    \includegraphics[width=\textwidth]{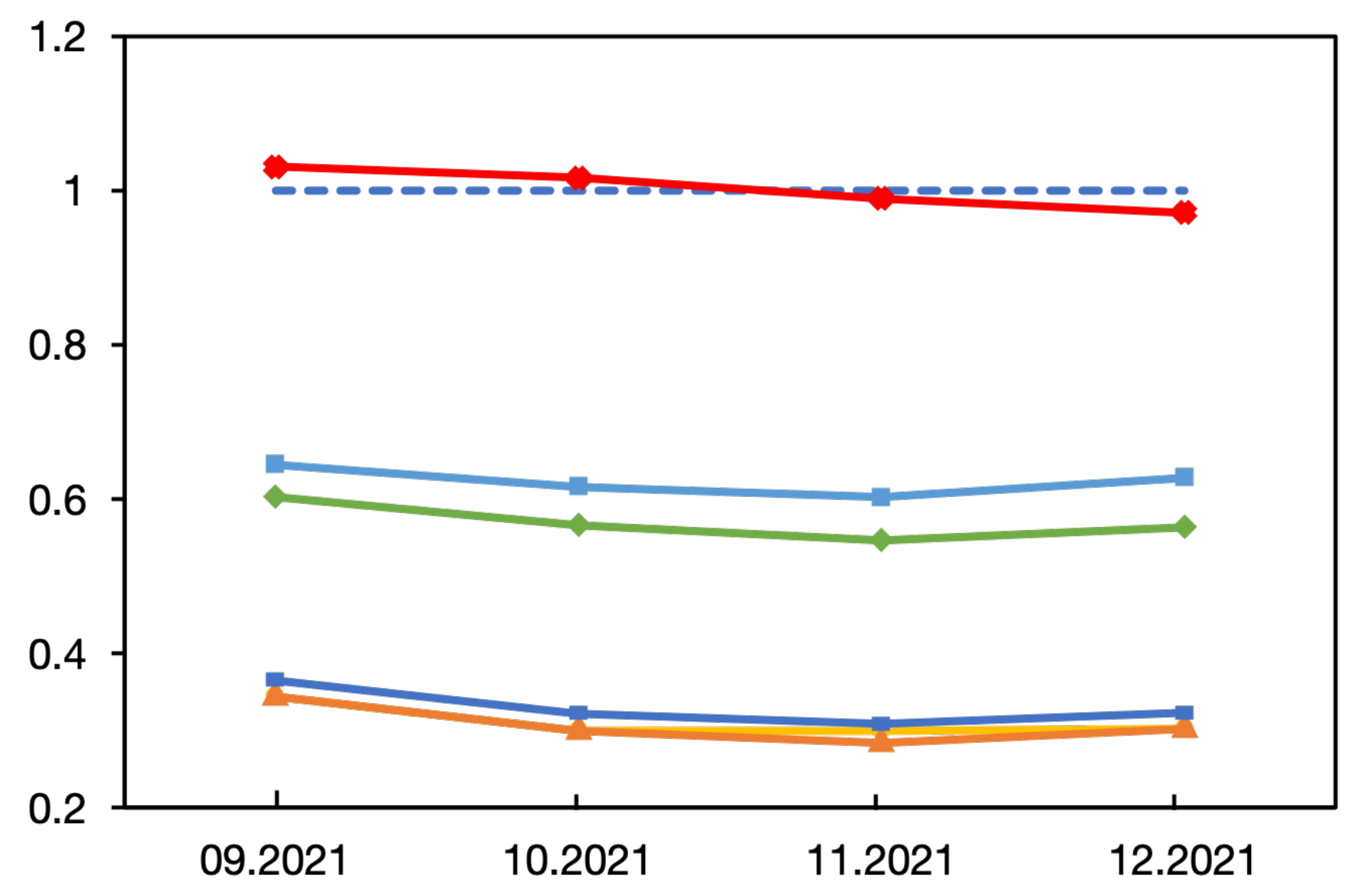}
    \caption{\textsc{TWiki-Diffsets}}
    \end{subfigure}
    \caption{Relative proper noun perplexity of \textsc{Full}, \textsc{Diff}, and \textsc{K-Adapter}, \textsc{LoRA}, \textsc{RecAdam} and \textsc{Mix-review} compared to \textsc{Initial} on \textsc{TWiki-Diffsets} and \textsc{Non-TWiki-Diffsets} for each month. Lower ratio indicates better performance. The performance of \textsc{Diff} (orange) and \textsc{RecAdam} (yellow) in (b) is is almost identical.}
    \label{fig:gpt2_instrinsic}
\vspace{-5mm}
\end{figure}

\begin{table*}[h!]
\caption{Zero-shot perplexity of LMs measured on \textsc{TWiki-Probes}. \textbf{Time} represents the average training time of a single update under the setting described in Section \ref{sec:settings}. The description of each baseline model is explained in Section \ref{sec:baselines}. Best performance is marked as \textbf{bold} while the second best is \underline{underlined}.}
    \resizebox{\textwidth}{!}{\begin{tabular}{cccccccccccccc}
    \toprule
    \multicolumn{1}{l}{} & \multicolumn{1}{l}{} & \multicolumn{3}{c}{TWiki-Probes-0809} & \multicolumn{3}{c}{TWiki-Probes-0910} & \multicolumn{3}{c}{TWiki-Probes-1011} &
    \multicolumn{3}{c}{TWiki-Probes-1112}
    \\ \cmidrule(lr){3-5} \cmidrule(lr){6-8} \cmidrule(lr){9-11} \cmidrule(lr){12-14}  & \textbf{Time} & \textbf{Un} & \textbf{C} & \textbf{Avg} & \textbf{Un} & \textbf{C} & \textbf{Avg} & \textbf{Un} & \textbf{C} & \textbf{Avg} & \textbf{Un} & \textbf{C} & \textbf{Avg} \\
    \midrule
    \textsc{Initial} & 0 hours & 386.16 & 364.82 & 375.49 & \underline{356.66} & 416.32 & 386.49 & 350.54 & 420.52 & 385.53 &  357.37 & 451.74 & 404.56\\ \midrule
    \textsc{Full} & $\sim$24 hours & 379.43 & 360.46 & 369.95 & 388.85 & 437.15 & 413.00 & \underline{337.34} & 383.06 & \underline{360.20} & 381.11 & 435.47 & 408.29\\
    \textsc{Diff} & $\sim$2.5 hours & 409.31 & 284.34 & 346.83 & 409.86 & \underline{336.55} & 373.21 & 465.20 & 367.72 & 416.46 & 391.77 & 365.07 & 378.42 \\ \midrule
    \textsc{RecAdam} & $\sim$4 hours & 358.10 & \textbf{253.07} & \textbf{305.59} & 376.12 & \textbf{306.64} & \underline{341.38} & 439.14 & \underline{338.17} & 388.66 & 400.56 & \underline{356.60} & 378.58\\
    \textsc{Mix-review} & $\sim$6 hours & \textbf{337.59} & \underline{274.91} & \underline{306.25} & 394.20 & 381.21 & 387.71 & 375.85 & 369.50 & 372.68 & \textbf{313.94} & \textbf{323.49} & \textbf{318.72} \\
     \textsc{LoRA} & $\sim$2 hours & 386.52 & 332.98 & 359.75 & 359.54 & 371.03 & 365.29 & 381.80 & 391.66 & 386.73 & 361.42 & 408.19 & 384.81 \\
    \textsc{K-Adapter} & $\sim$2 hours & \underline{340.47} & 297.39 & 318.93 & \textbf{326.53} & 338.16 & \textbf{332.35} & \textbf{325.11} & \textbf{332.61} & \textbf{328.86} & \underline{333.53} & 374.67 & \underline{354.10} \\
    \bottomrule
    \end{tabular}}
\label{table:gpt2_zero-shot}
\vspace{-5mm}
\end{table*}

We first perform intrinsic evaluation by measuring the perplexity of the baseline models on their training corpora. For each month, we measure the model's perplexity on \textsc{TWiki-Diffsets} and \textsc{Non-TWiki-Diffsets}, where the latter refers to the subset of the month's entire Wikipedia snapshot that does not include the data from \textsc{TWiki-Diffsets}. We sample 10,000 input instances from each subset with a fixed length of 512 and measure the perplexity on proper noun tokens determined by a Part-of-Speech (POS) tagger~\citep{spacy2} as in \citep{lazaridou2021mind}, which can be considered as a proxy for tokens containing factual knowledge. Therefore, the result on \textsc{Non-TWiki-Diffsets} is meant to indicate the performance on unchanged knowledge, while the result on \textsc{TWiki-Diffsets} corresponds to updated and new knowledge. Figure \ref{fig:gpt2_instrinsic} shows the relative perplexity of each baseline method compared to \textsc{Initial} (i.e., dividing each model by \textsc{Initial}, and thus the lower, the better).

Results on \textsc{Non-TWiki-Diffsets} show that the relative perplexity of \textsc{Diff} increases while that of \textsc{Full} remains constant as time goes on, which implies that forgetting occurs when the LM is trained with \textsc{TWiki-Diffsets}. The relative perplexities of continual learning methods increase less rapidly than \textsc{Diff}, which means that applying continual learning mitigates catastrophic forgetting. \textsc{Mix-review}, especially, shows the least amount of forgetting among the continual learning methods, which indicates that training on the past corpus is effective in retaining performance on the previous training corpora in terms of perplexity.

On the other hand, the results on \textsc{TWiki-Diffsets} show the opposite trend: the relative perplexity of \textsc{Diff} is much lower than \textsc{Full}. One thing to note is that the perplexity of \textsc{Full} is very similar to that of \textsc{Initial} on \textsc{TWiki-Diffsets}, which suggests that updating LMs on entire Wikipedia snapshots hinders the effective learning of \textit{changed} data compared to \textsc{Diff}, despite both having seen the same instances of \textsc{TWiki-Diffsets} during training for the same number of iterations. Among continual learning methods, \textsc{K-Adapter} and \textsc{LoRA} shows higher overall perplexities than \textsc{Diff} while \textsc{Mix-review} and \textsc{RecAdam} shows similar perplexity.
\subsection{Extrinsic Evaluation on \textsc{TWiki-Probes}}
\label{sec:extrinsic}

\begin{figure}[t!]  
    \centering
    \includegraphics[width=0.9\columnwidth]{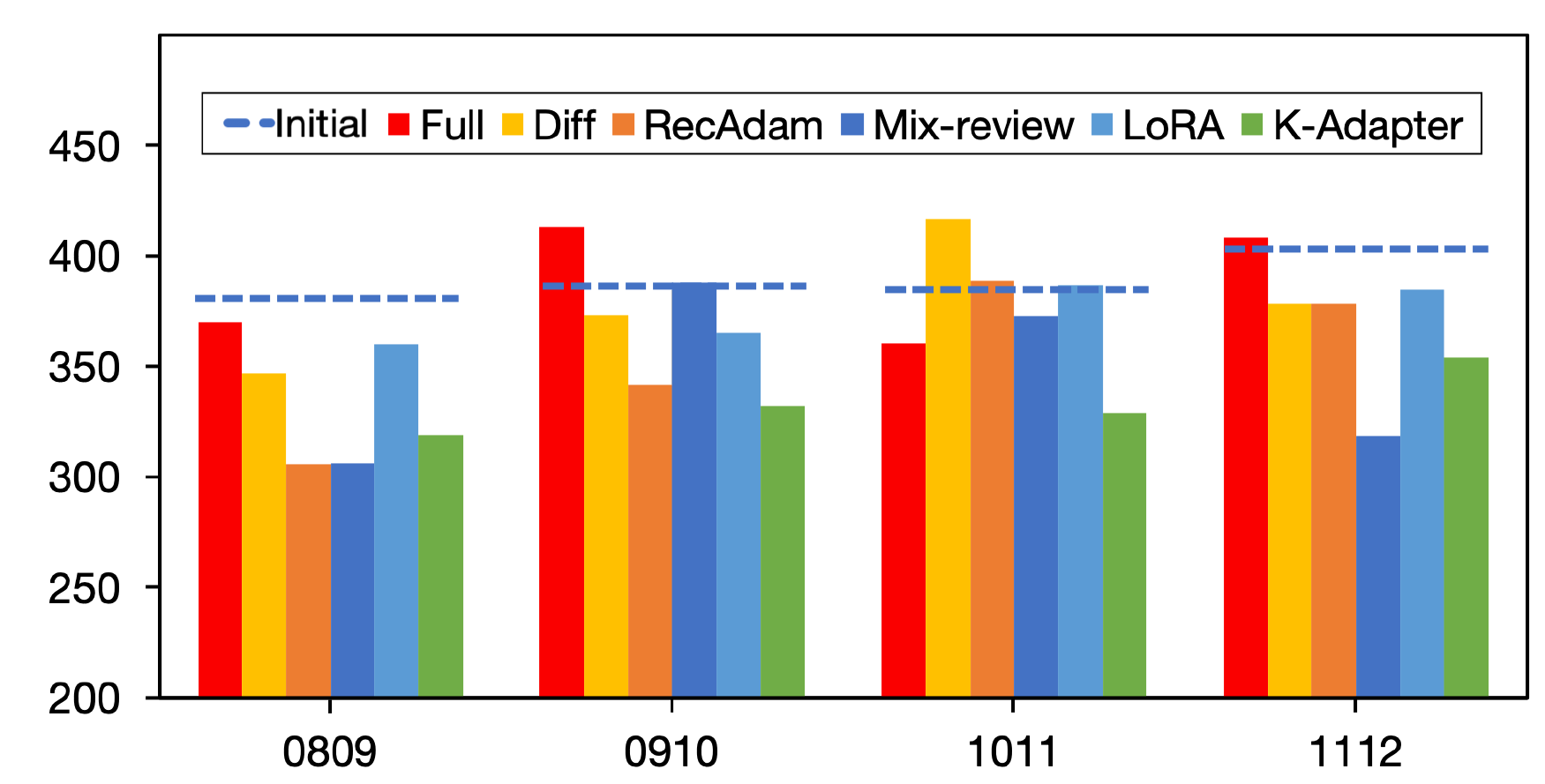}
    \caption{Average overall perplexity of \textsc{TWiki-Probes}. We average the perplexities of \textsc{Unchanged} and \textsc{Changed} with equal importance placed on stability and plasticity. The x-axis depicts the two-month intervals. A lower score indicates better performance.}
    \label{fig:overall}
\vspace{-5mm}
\end{figure}

\begin{figure*}[ht!]
\centering
    \begin{subfigure}[b]{0.27\textwidth}
    \includegraphics[width=\textwidth]{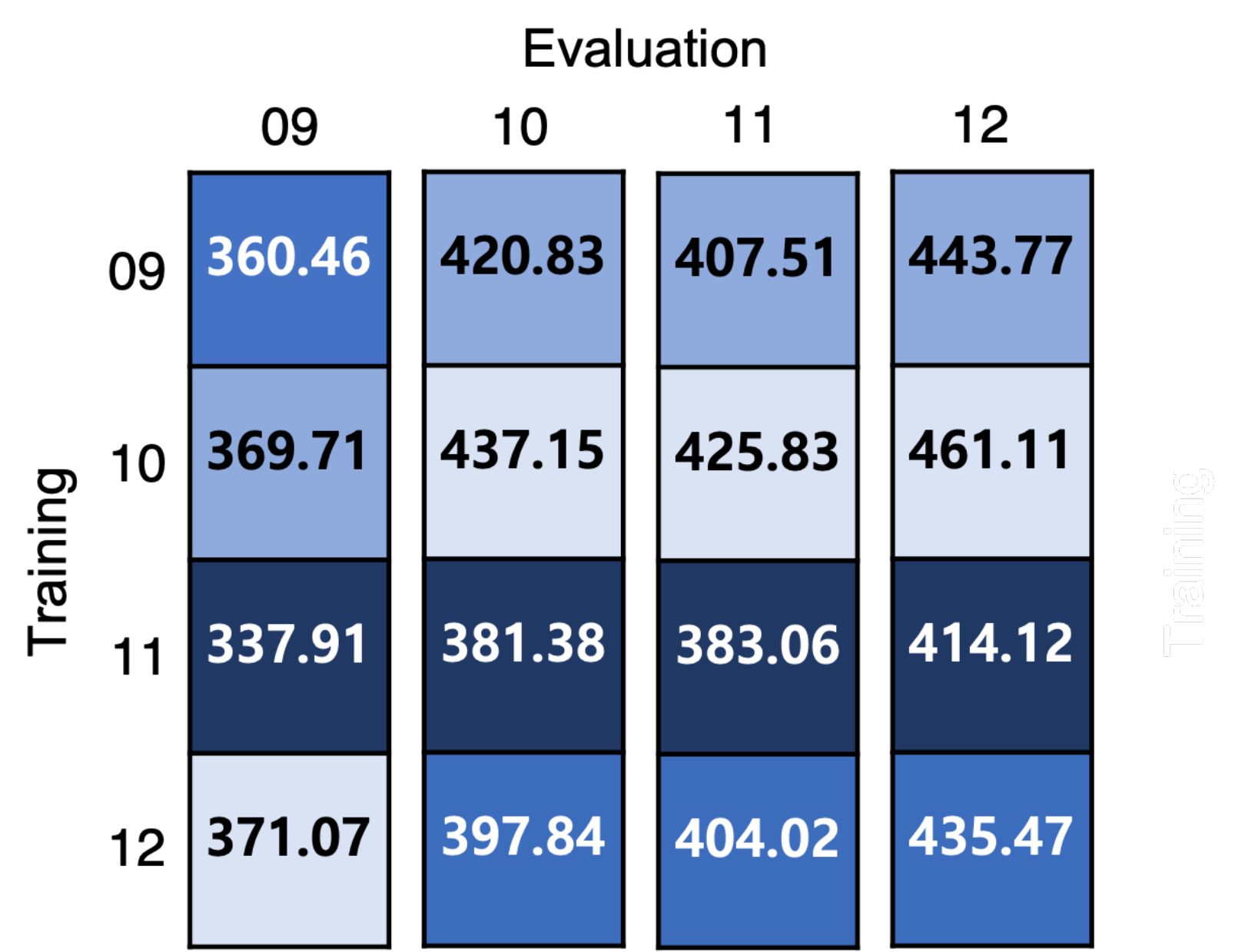}
    \caption{\textsc{Full}}
    \end{subfigure}
    \begin{subfigure}[b]{0.27\textwidth}
    \includegraphics[width=\textwidth]{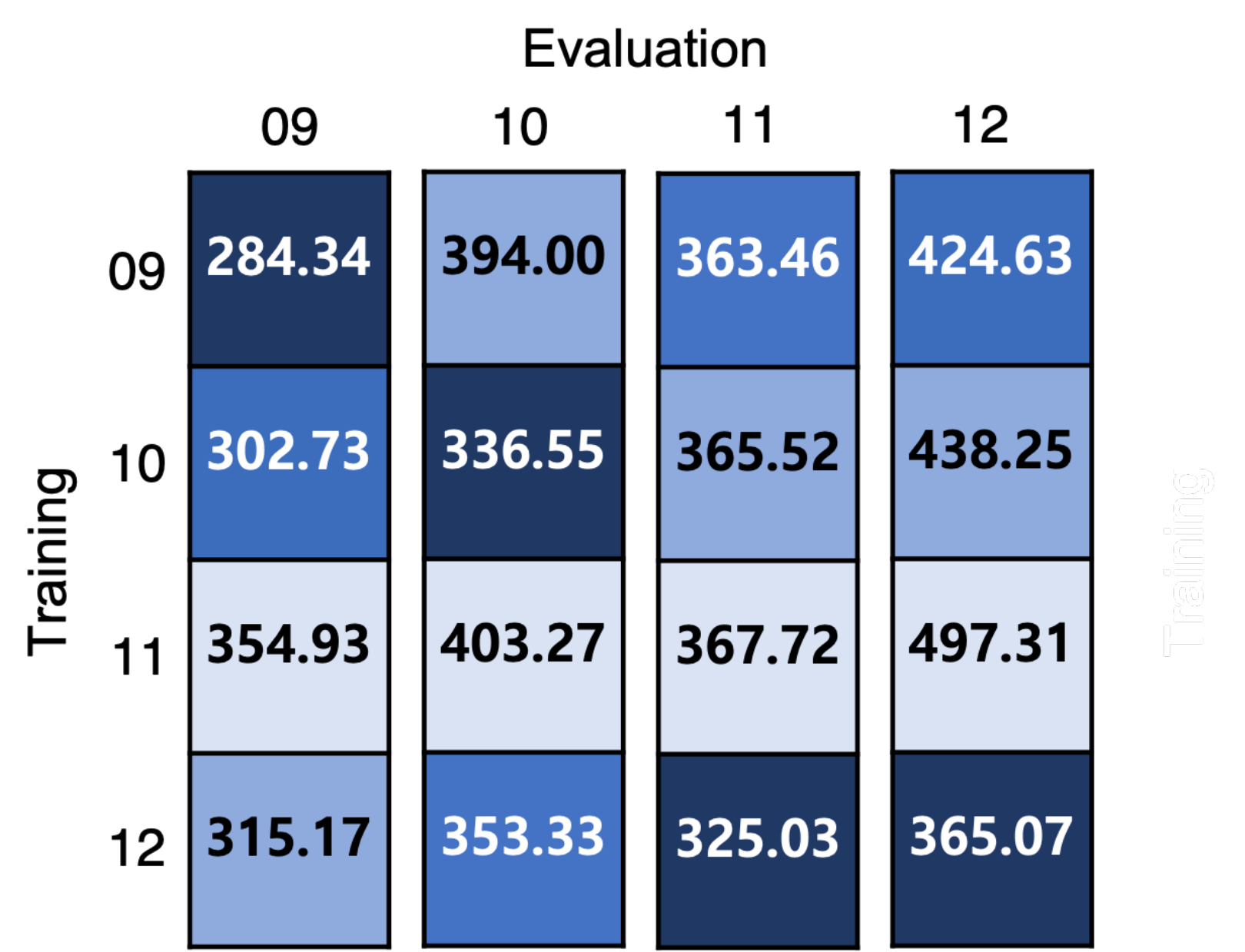}
    \caption{\textsc{Diff}}
    \end{subfigure}
    \begin{subfigure}[b]{0.27\textwidth}
    \includegraphics[width=\textwidth]{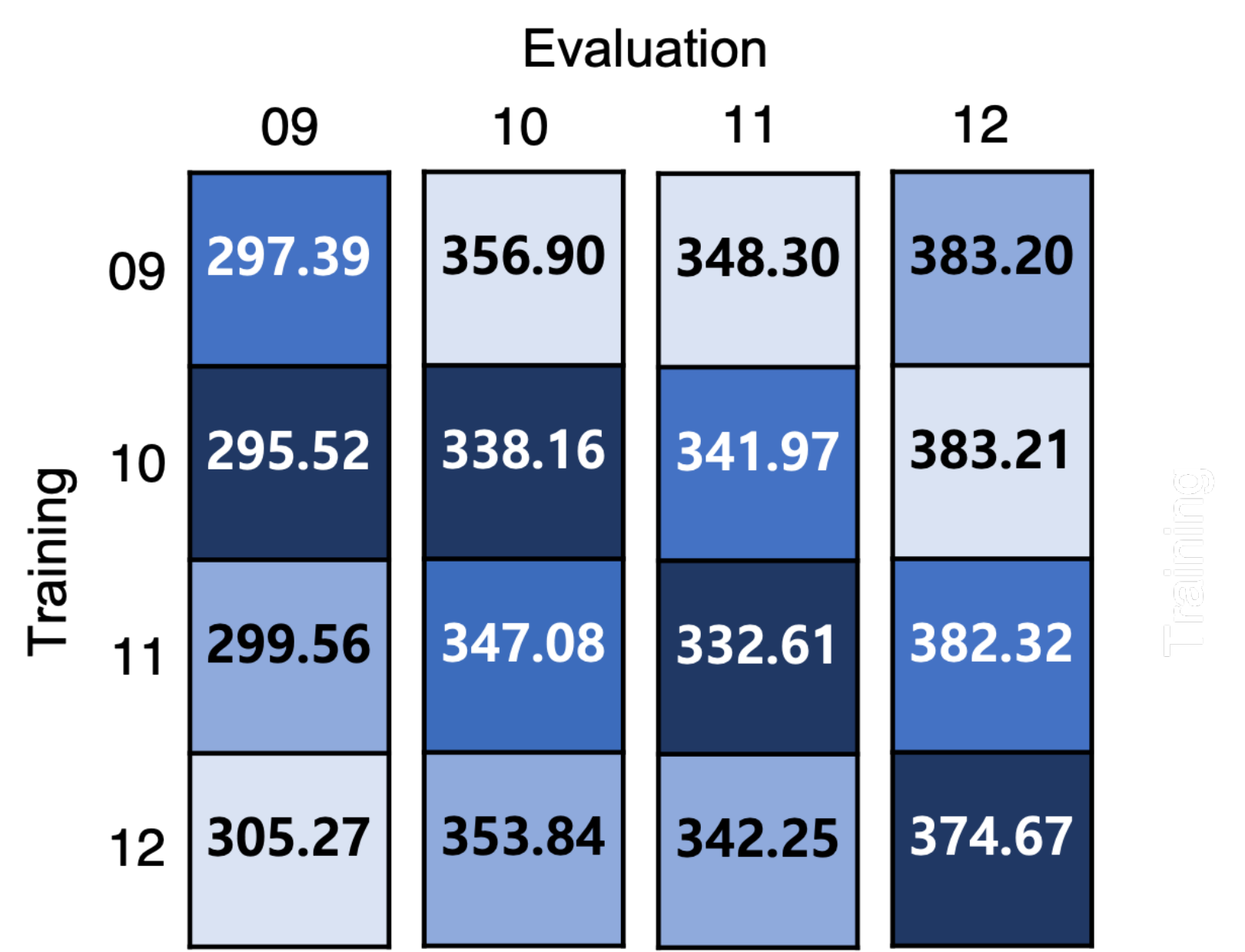}
    \caption{\textsc{K-Adapter}}
    \end{subfigure}
\caption{The zero-shot perplexity of the LMs updated and evaluated on various time intervals of \textsc{Changed} of \textsc{TWiki-Probes}, showing the effect of temporal misalignment. The better the results, the darker the performance is colored. The vertical axis represents the Trai.}
\label{fig:temp_misalign}
\vspace{-5mm}
\end{figure*}

Performing only intrinsic evaluation on the training corpora is not sufficient because the intrinsic evaluation itself only tests the capability of the LMs for memorization~\citep{mccoy2021much}. Through extrinsic evaluation with \textsc{TWiki-Probes} (Section~\ref{sec:eval}), we specifically focus on evaluating \textit{factual} knowledge of the LMs from each update. Placing equal importance on \textit{stability} (\textsc{Unchanged}) and \textit{plasticity} (\textsc{Changed}), we show the average of the perplexities of \textsc{Unchanged} and \textsc{Changed} as well as individual perplexities in Table \ref{table:gpt2_zero-shot}, and show a bar graph of the average perplexities in Figure \ref{fig:overall}\footnote{The perplexity of \textsc{Unchanged} and \textsc{Changed} were each calculated by measuring the average perplexity of generating each factual instances.}. 

As shown in Table \ref{table:gpt2_zero-shot}, \textsc{Diff} and all continual learning methods show better overall performance on \textsc{Changed} factual instances than \textsc{Initial} in all months, bolstering the results from the intrinsic evaluation. For \textsc{Unchanged}, however, \textsc{Diff} suffers from \textit{catastrophic forgetting}, showing consistent performance degradation as the number of updates increases. In contrast, continual learning methods effectively mitigate much of the catastrophic forgetting during temporal language modeling, resulting in lower perplexity on \textsc{Unchanged}, except \textsc{RecAdam} which performs worse as the number of updates increases. \textsc{K-Adapter}, especially, shows surprising results on \textsc{Unchanged}, outperforming even \textsc{Full} throughout all of the months. Moreover, all continual learning methods surpass or are on par with \textsc{Diff} on \textsc{Changed} factual instances, showing that ability to learn new knowledge (plasticity) is not sacrificed to preserve previous knowledge (stability).

Moreover, as shown in the average perplexity column of Table \ref{table:gpt2_zero-shot} and Figure \ref{fig:overall}, \textsc{K-Adapter} shows the most robust performance throughout the time periods. It is important to note that \textsc{K-Adapter} is around 12 times more computationally efficient than \textsc{Full} in terms of total training time, under the same computational constraint. \textsc{Diff} also outperforms \textsc{Full} in all months but 1011, showing that temporal language modeling itself is an effective approach for overall stability-plasticity trade-off. 

We note that, as also shown in previous works~\citep{lazaridou2021mind}, results in Table \ref{table:gpt2_zero-shot} present an overall high perplexity (>200) because the sentences in \textsc{TWiki-Probes} are not natural sentences; they are factual phrases \textit{synthetically} generated from a naive concatenation of \texttt{Subject}, \texttt{Relation}, and \texttt{Object}. We address this issue via \textit{light-tuning} in Appendix \ref{appen:f1}.

\paragraph{Effect of Temporal Misalignment}
We quantify the effect of temporal misalignment on each method by training the LMs and evaluating their zero-shot perplexity on \textsc{Changed} instances of \textsc{TWiki-Probes} with various time intervals of training and evaluation. Among continual learning methods, we select \textsc{K-Adapter} since it shows the most robust performance for extrinsic evaluation across all time periods. As shown in Figure \ref{fig:temp_misalign}, \textsc{Full} method is mostly influenced by the number of training updates and not much by whether there is temporal alignment. Since \textsc{Full} is continuously pretrained on the entire Wikipedia corpus in each month, it would have likely seen the data containing \textsc{Changed} factual instances multiple times, leading to lower perplexity as training steps increases.\footnote{Although directly training \textsc{Initial}  on the whole Wikipedia corpus of a specific month
can be an alternative, we exclude it here because it would only learn the knowledge of the specific month and thus inappropriate for a truly ever-evolving setting.} 
For \textsc{Diff} and \textsc{K-Adapter}, there is a general trend of strong performance when there is temporal alignment (diagonal entries),
outperforming \textsc{Full} with much fewer global training steps. It is important to note that \textsc{K-Adapter} shows robustness against temporal misalignment, i.e., the perplexity does not increase much even when the training and evaluation months do not match, compared to \textsc{Diff} which suffers from a more severe perplexity spike.


\section{Conclusion}
In this paper, we provide answers to the four proposed questions in Section \ref{sec:introduction}. (1) \textit{How can we train ever-evolving LMs efficiently and automate the evaluation of each update?} We introduce \textsc{TemporalWiki}, a lifelong benchmark that can be used for training and evaluating ever-evolving LMs in an automated manner. It consists of \textsc{TWiki-Diffsets} as the training corpora for temporal language modeling and \textsc{TWiki-Probes} as the evaluation datasets for measuring the stability-plasticity trade-off. (2) \textit{How does updating LMs only on new and updated data from Wikipedia compare to updating LMs on entire Wikipedia snapshots, especially in scenarios with multiple updates?} Through experiments on \textsc{TemporalWiki}, we show that updating LMs on \textsc{TWiki-Diffsets} leads to better acquisition of \textit{new} and \textit{updated} knowledge than updating on entire Wikipedia snapshots with much less computational cost (12 times less). 
(3) \textit{How serious is catastrophic forgetting when LMs are updated only on new and updated data?} We observe that temporal language modeling is a challenging problem, especially as the number of LM updates increases. However, results still show an overall enhancement in terms of stability-plasticity compared to updating with entire Wikipedia snapshots, showing that temporal language modeling is an effective alternative. 
(4) \textit{How can we mitigate catastrophic forgetting?} We find that continual learning methods (regularization, rehearsal, and parameter-expansion) for large language model training effectively mitigates forgetting and shows robust performance in terms of enhancing the overall trade-off between stability and plasticity on \textsc{TWiki-Probes}.

\section{Limitations}
\label{sec:limit}
As mentioned at the beginning of this Section, each Wikipedia and Wikidata update does not ensure an actual update of \textit{real-world} knowledge. For example, an addition of a new Wikipedia page does not necessarily mean that all the information on the new page is \textit{new} world knowledge. Likewise, \textit{existing} factual knowledge may be added to Wikidata because Wikipedia and Wikidata do not cover all of the world knowledge and may have some missing information about the world.
Moreover, one aspect that is not covered in this work is \textit{knowledge deletion}. While maintaining Wikipedia and Wikidata, volunteer editors not only update or add new information but also \textit{delete} information that is incorrect or misinformed. As removing the misinformation and bias stored in LMs is an important issue and necessary for truly ever-evolving LMs, future work should address this aspect utilizing \textit{deleted} information from general knowledge sources such as Wikipedia.

\section*{Acknowledgements}
This work was supported by Institute of Information \& communications Technology Planning \& Evaluation (IITP) grants funded by the Korea government (MSIT) (No.2022-0-00113, Developing a Sustainable Collaborative Multi-modal Lifelong Learning Framework, 80\%; No.2019-0-00075, Artificial Intelligence Graduate School Program (KAIST), 10\%; No.2021-0-02068, Artificial Intelligence Innovation Hub, 10\%).

\bibliography{anthology,custom}

\begin{thebibliography}{42}
\expandafter\ifx\csname natexlab\endcsname\relax\def\natexlab#1{#1}\fi

\bibitem[{Agarwal and Nenkova(2021)}]{agarwal2021temporal}
Oshin Agarwal and Ani Nenkova. 2021.
\newblock Temporal effects on pre-trained models for language processing tasks.
\newblock \emph{arXiv preprint arXiv:2111.12790}.

\bibitem[{Borgeaud et~al.(2021)Borgeaud, Mensch, Hoffmann, Cai, Rutherford,
  Millican, Driessche, Lespiau, Damoc, Clark et~al.}]{borgeaud2021improving}
Sebastian Borgeaud, Arthur Mensch, Jordan Hoffmann, Trevor Cai, Eliza
  Rutherford, Katie Millican, George van~den Driessche, Jean-Baptiste Lespiau,
  Bogdan Damoc, Aidan Clark, et~al. 2021.
\newblock Improving language models by retrieving from trillions of tokens.
\newblock \emph{arXiv preprint arXiv:2112.04426}.

\bibitem[{Brown et~al.(2020)Brown, Mann, Ryder, Subbiah, Kaplan, Dhariwal,
  Neelakantan, Shyam, Sastry, Askell et~al.}]{brown2020language}
Tom~B Brown, Benjamin Mann, Nick Ryder, Melanie Subbiah, Jared Kaplan, Prafulla
  Dhariwal, Arvind Neelakantan, Pranav Shyam, Girish Sastry, Amanda Askell,
  et~al. 2020.
\newblock Language models are few-shot learners.
\newblock In \emph{NeurIPS}.

\bibitem[{Chen et~al.(2020)Chen, Hou, Cui, Che, Liu, and Yu}]{chen2020recall}
Sanyuan Chen, Yutai Hou, Yiming Cui, Wanxiang Che, Ting Liu, and Xiangzhan Yu.
  2020.
\newblock \href {https://doi.org/10.18653/v1/2020.emnlp-main.634} {Recall and
  learn: Fine-tuning deep pretrained language models with less forgetting}.
\newblock In \emph{Proceedings of the 2020 Conference on Empirical Methods in
  Natural Language Processing (EMNLP)}, pages 7870--7881, Online. Association
  for Computational Linguistics.

\bibitem[{Chen et~al.(2021)Chen, Wang, and Wang}]{chen2021dataset}
Wenhu Chen, Xinyi Wang, and William~Yang Wang. 2021.
\newblock A dataset for answering time-sensitive questions.
\newblock In \emph{NeurIPS}.

\bibitem[{Dhingra et~al.(2022)Dhingra, Cole, Eisenschlos, Gillick, Eisenstein,
  and Cohen}]{dhingra2021time}
Bhuwan Dhingra, Jeremy~R. Cole, Julian~Martin Eisenschlos, Daniel Gillick,
  Jacob Eisenstein, and William~W. Cohen. 2022.
\newblock \href {https://doi.org/10.1162/tacl_a_00459} {Time-aware language
  models as temporal knowledge bases}.
\newblock \emph{Transactions of the Association for Computational Linguistics},
  10:257--273.

\bibitem[{Dinan et~al.(2019)Dinan, Roller, Shuster, Fan, Auli, and
  Weston}]{dinan2018wizard}
Emily Dinan, Stephen Roller, Kurt Shuster, Angela Fan, Michael Auli, and Jason
  Weston. 2019.
\newblock Wizard of wikipedia: Knowledge-powered conversational agents.
\newblock In \emph{ICLR}.

\bibitem[{Guu et~al.(2020)Guu, Lee, Tung, Pasupat, and Chang}]{guu2020realm}
Kelvin Guu, Kenton Lee, Zora Tung, Panupong Pasupat, and Ming-Wei Chang. 2020.
\newblock Realm: Retrieval-augmented language model pre-training.
\newblock In \emph{ICML}.

\bibitem[{He et~al.(2021)He, Liu, Cho, Ott, Liu, Glass, and
  Peng}]{he2021analyzing}
Tianxing He, Jun Liu, Kyunghyun Cho, Myle Ott, Bing Liu, James Glass, and
  Fuchun Peng. 2021.
\newblock \href {https://doi.org/10.18653/v1/2021.eacl-main.95} {Analyzing the
  forgetting problem in pretrain-finetuning of open-domain dialogue response
  models}.
\newblock In \emph{Proceedings of the 16th Conference of the European Chapter
  of the Association for Computational Linguistics: Main Volume}, pages
  1121--1133, Online. Association for Computational Linguistics.

\bibitem[{Hombaiah et~al.(2021)Hombaiah, Chen, Zhang, Bendersky, and
  Najork}]{hombaiah2021dynamic}
Spurthi~Amba Hombaiah, Tao Chen, Mingyang Zhang, Michael Bendersky, and Marc
  Najork. 2021.
\newblock Dynamic language models for continuously evolving content.
\newblock In \emph{KDD}.

\bibitem[{Honnibal and Montani(2017)}]{spacy2}
Matthew Honnibal and Ines Montani. 2017.
\newblock {spaCy 2}: Natural language understanding with {B}loom embeddings,
  convolutional neural networks and incremental parsing.
\newblock To appear.

\bibitem[{Hu et~al.(2022)Hu, Shen, Wallis, Allen-Zhu, Li, Wang, and
  Chen}]{hu2021lora}
Edward~J Hu, Yelong Shen, Phillip Wallis, Zeyuan Allen-Zhu, Yuanzhi Li, Shean
  Wang, and Weizhu Chen. 2022.
\newblock Lora: Low-rank adaptation of large language models.
\newblock \emph{ICLR}.

\bibitem[{Jang et~al.(2022)Jang, Ye, Yang, Shin, Han, Kim, Choi, and
  Seo}]{jang2021towards}
Joel Jang, Seonghyeon Ye, Sohee Yang, Joongbo Shin, Janghoon Han, Gyeonghun
  Kim, Stanley~Jungkyu Choi, and Minjoon Seo. 2022.
\newblock Towards continual knowledge learning of language models.
\newblock In \emph{ICLR}.

\bibitem[{Jin et~al.(2022)Jin, Zhang, Zhu, Xiao, Li, Wei, Arnold, and
  Ren}]{jin2021lifelong}
Xisen Jin, Dejiao Zhang, Henghui Zhu, Wei Xiao, Shang-Wen Li, Xiaokai Wei,
  Andrew Arnold, and Xiang Ren. 2022.
\newblock \href {https://doi.org/10.18653/v1/2022.bigscience-1.1} {Lifelong
  pretraining: Continually adapting language models to emerging corpora}.
\newblock In \emph{Proceedings of BigScience Episode {\#}5 -- Workshop on
  Challenges {\&} Perspectives in Creating Large Language Models}, pages 1--16,
  virtual+Dublin. Association for Computational Linguistics.

\bibitem[{Kirkpatrick et~al.(2017)Kirkpatrick, Pascanu, Rabinowitz, Veness,
  Desjardins, Rusu, Milan, Quan, Ramalho, Grabska-Barwinska
  et~al.}]{kirkpatrick2017overcoming}
James Kirkpatrick, Razvan Pascanu, Neil Rabinowitz, Joel Veness, Guillaume
  Desjardins, Andrei~A Rusu, Kieran Milan, John Quan, Tiago Ramalho, Agnieszka
  Grabska-Barwinska, et~al. 2017.
\newblock Overcoming catastrophic forgetting in neural networks.
\newblock \emph{Proceedings of the national academy of sciences},
  114(13):3521--3526.

\bibitem[{Kwiatkowski et~al.(2019)Kwiatkowski, Palomaki, Redfield, Collins,
  Parikh, Alberti, Epstein, Polosukhin, Devlin, Lee, Toutanova, Jones, Kelcey,
  Chang, Dai, Uszkoreit, Le, and Petrov}]{kwiatkowski2019natural}
Tom Kwiatkowski, Jennimaria Palomaki, Olivia Redfield, Michael Collins, Ankur
  Parikh, Chris Alberti, Danielle Epstein, Illia Polosukhin, Jacob Devlin,
  Kenton Lee, Kristina Toutanova, Llion Jones, Matthew Kelcey, Ming-Wei Chang,
  Andrew~M. Dai, Jakob Uszkoreit, Quoc Le, and Slav Petrov. 2019.
\newblock \href {https://doi.org/10.1162/tacl_a_00276} {Natural questions: A
  benchmark for question answering research}.
\newblock \emph{Transactions of the Association for Computational Linguistics},
  7:452--466.

\bibitem[{Lazaridou et~al.(2021)Lazaridou, Kuncoro, Gribovskaya, Agrawal,
  Liska, Terzi, Gimenez, de~Masson~d'Autume, Kocisky, Ruder
  et~al.}]{lazaridou2021mind}
Angeliki Lazaridou, Adhi Kuncoro, Elena Gribovskaya, Devang Agrawal, Adam
  Liska, Tayfun Terzi, Mai Gimenez, Cyprien de~Masson~d'Autume, Tomas Kocisky,
  Sebastian Ruder, et~al. 2021.
\newblock Mind the gap: Assessing temporal generalization in neural language
  models.
\newblock In \emph{NeurIPS}.

\bibitem[{Lewis et~al.(2020)Lewis, Perez, Piktus, Petroni, Karpukhin, Goyal,
  K{\"u}ttler, Lewis, Yih, Rockt{\"a}schel et~al.}]{lewis2020retrieval}
Patrick Lewis, Ethan Perez, Aleksandra Piktus, Fabio Petroni, Vladimir
  Karpukhin, Naman Goyal, Heinrich K{\"u}ttler, Mike Lewis, Wen-tau Yih, Tim
  Rockt{\"a}schel, et~al. 2020.
\newblock Retrieval-augmented generation for knowledge-intensive nlp tasks.
\newblock In \emph{NeurIPS}.

\bibitem[{Lewis et~al.(2021)Lewis, Stenetorp, and
  Riedel}]{lewis-etal-2021-question}
Patrick Lewis, Pontus Stenetorp, and Sebastian Riedel. 2021.
\newblock \href {https://doi.org/10.18653/v1/2021.eacl-main.86} {Question and
  answer test-train overlap in open-domain question answering datasets}.
\newblock In \emph{Proceedings of the 16th Conference of the European Chapter
  of the Association for Computational Linguistics: Main Volume}, pages
  1000--1008, Online. Association for Computational Linguistics.

\bibitem[{Logan~IV et~al.(2021)Logan~IV, Passos, Singh, and
  Chang}]{logan2021fruit}
Robert~L Logan~IV, Alexandre Passos, Sameer Singh, and Ming-Wei Chang. 2021.
\newblock Fruit: Faithfully reflecting updated information in text.
\newblock \emph{arXiv preprint arXiv:2112.08634}.

\bibitem[{Longpre et~al.(2021)Longpre, Perisetla, Chen, Ramesh, DuBois, and
  Singh}]{longpre-etal-2021-entity}
Shayne Longpre, Kartik Perisetla, Anthony Chen, Nikhil Ramesh, Chris DuBois,
  and Sameer Singh. 2021.
\newblock \href {https://doi.org/10.18653/v1/2021.emnlp-main.565} {Entity-based
  knowledge conflicts in question answering}.
\newblock In \emph{Proceedings of the 2021 Conference on Empirical Methods in
  Natural Language Processing}, pages 7052--7063, Online and Punta Cana,
  Dominican Republic. Association for Computational Linguistics.

\bibitem[{Loureiro et~al.(2022)Loureiro, Barbieri, Neves, Espinosa~Anke, and
  Camacho-collados}]{loureiro2022timelms}
Daniel Loureiro, Francesco Barbieri, Leonardo Neves, Luis Espinosa~Anke, and
  Jose Camacho-collados. 2022.
\newblock \href {https://doi.org/10.18653/v1/2022.acl-demo.25} {{T}ime{LM}s:
  Diachronic language models from {T}witter}.
\newblock In \emph{Proceedings of the 60th Annual Meeting of the Association
  for Computational Linguistics: System Demonstrations}, pages 251--260,
  Dublin, Ireland. Association for Computational Linguistics.

\bibitem[{Luu et~al.(2021)Luu, Khashabi, Gururangan, Mandyam, and
  Smith}]{luu2021time}
Kelvin Luu, Daniel Khashabi, Suchin Gururangan, Karishma Mandyam, and Noah~A
  Smith. 2021.
\newblock Time waits for no one! analysis and challenges of temporal
  misalignment.
\newblock \emph{arXiv preprint arXiv:2111.07408}.

\bibitem[{McCloskey and Cohen(1989)}]{mccloskey1989catastrophic}
Michael McCloskey and Neal~J Cohen. 1989.
\newblock Catastrophic interference in connectionist networks: The sequential
  learning problem.
\newblock \emph{Psychology of learning and motivation}.

\bibitem[{McCoy et~al.(2021)McCoy, Smolensky, Linzen, Gao, and
  Celikyilmaz}]{mccoy2021much}
R~Thomas McCoy, Paul Smolensky, Tal Linzen, Jianfeng Gao, and Asli Celikyilmaz.
  2021.
\newblock How much do language models copy from their training data? evaluating
  linguistic novelty in text generation using raven.
\newblock \emph{arXiv preprint arXiv:2111.09509}.

\bibitem[{Mermillod et~al.(2013)Mermillod, Bugaiska, and
  Bonin}]{mermillod2013stability}
Martial Mermillod, Aur{\'e}lia Bugaiska, and Patrick Bonin. 2013.
\newblock The stability-plasticity dilemma: Investigating the continuum from
  catastrophic forgetting to age-limited learning effects.
\newblock \emph{Frontiers in Psychology}.

\bibitem[{Osborne et~al.(2014)Osborne, Lall, and
  Van~Durme}]{osborne-etal-2014-exponential}
Miles Osborne, Ashwin Lall, and Benjamin Van~Durme. 2014.
\newblock \href {https://doi.org/10.3115/v1/P14-2112} {Exponential reservoir
  sampling for streaming language models}.
\newblock In \emph{Proceedings of the 52nd Annual Meeting of the Association
  for Computational Linguistics (Volume 2: Short Papers)}, pages 687--692,
  Baltimore, Maryland. Association for Computational Linguistics.

\bibitem[{Petroni et~al.(2021)Petroni, Piktus, Fan, Lewis, Yazdani, De~Cao,
  Thorne, Jernite, Karpukhin, Maillard, Plachouras, Rockt{\"a}schel, and
  Riedel}]{petroni2020kilt}
Fabio Petroni, Aleksandra Piktus, Angela Fan, Patrick Lewis, Majid Yazdani,
  Nicola De~Cao, James Thorne, Yacine Jernite, Vladimir Karpukhin, Jean
  Maillard, Vassilis Plachouras, Tim Rockt{\"a}schel, and Sebastian Riedel.
  2021.
\newblock \href {https://doi.org/10.18653/v1/2021.naacl-main.200} {{KILT}: a
  benchmark for knowledge intensive language tasks}.
\newblock In \emph{Proceedings of the 2021 Conference of the North American
  Chapter of the Association for Computational Linguistics: Human Language
  Technologies}, pages 2523--2544, Online. Association for Computational
  Linguistics.

\bibitem[{Petroni et~al.(2019)Petroni, Rockt{\"a}schel, Riedel, Lewis, Bakhtin,
  Wu, and Miller}]{petroni2019language}
Fabio Petroni, Tim Rockt{\"a}schel, Sebastian Riedel, Patrick Lewis, Anton
  Bakhtin, Yuxiang Wu, and Alexander Miller. 2019.
\newblock \href {https://doi.org/10.18653/v1/D19-1250} {Language models as
  knowledge bases?}
\newblock In \emph{Proceedings of the 2019 Conference on Empirical Methods in
  Natural Language Processing and the 9th International Joint Conference on
  Natural Language Processing (EMNLP-IJCNLP)}, pages 2463--2473, Hong Kong,
  China. Association for Computational Linguistics.

\bibitem[{Piktus et~al.(2021)Piktus, Petroni, Karpukhin, Okhonko, Broscheit,
  Izacard, Lewis, O{\u{g}}uz, Grave, Yih et~al.}]{piktus2021web}
Aleksandra Piktus, Fabio Petroni, Vladimir Karpukhin, Dmytro Okhonko, Samuel
  Broscheit, Gautier Izacard, Patrick Lewis, Barlas O{\u{g}}uz, Edouard Grave,
  Wen-tau Yih, et~al. 2021.
\newblock The web is your oyster--knowledge-intensive nlp against a very large
  web corpus.
\newblock \emph{arXiv preprint arXiv:2112.09924}.

\bibitem[{Radford et~al.(2019)Radford, Wu, Child, Luan, Amodei, Sutskever
  et~al.}]{radford2019language}
Alec Radford, Jeffrey Wu, Rewon Child, David Luan, Dario Amodei, Ilya
  Sutskever, et~al. 2019.
\newblock Language models are unsupervised multitask learners.
\newblock \emph{OpenAI blog}.

\bibitem[{Raffel et~al.(2019)Raffel, Shazeer, Roberts, Lee, Narang, Matena,
  Zhou, Li, and Liu}]{raffel2019exploring}
Colin Raffel, Noam Shazeer, Adam Roberts, Katherine Lee, Sharan Narang, Michael
  Matena, Yanqi Zhou, Wei Li, and Peter~J Liu. 2019.
\newblock Exploring the limits of transfer learning with a unified text-to-text
  transformer.
\newblock \emph{JMLR}.

\bibitem[{Roberts et~al.(2020)Roberts, Raffel, and Shazeer}]{roberts2020much}
Adam Roberts, Colin Raffel, and Noam Shazeer. 2020.
\newblock \href {https://doi.org/10.18653/v1/2020.emnlp-main.437} {How much
  knowledge can you pack into the parameters of a language model?}
\newblock In \emph{Proceedings of the 2020 Conference on Empirical Methods in
  Natural Language Processing (EMNLP)}, pages 5418--5426, Online. Association
  for Computational Linguistics.

\bibitem[{Rosin et~al.(2021)Rosin, Guy, and Radinsky}]{rosin2021time}
Guy~D Rosin, Ido Guy, and Kira Radinsky. 2021.
\newblock Time masking for temporal language models.
\newblock \emph{arXiv preprint arXiv:2110.06366}.

\bibitem[{R{\"o}ttger and Pierrehumbert(2021)}]{rottger2021temporal}
Paul R{\"o}ttger and Janet Pierrehumbert. 2021.
\newblock \href {https://doi.org/10.18653/v1/2021.findings-emnlp.206} {Temporal
  adaptation of {BERT} and performance on downstream document classification:
  Insights from social media}.
\newblock In \emph{Findings of the Association for Computational Linguistics:
  EMNLP 2021}, pages 2400--2412, Punta Cana, Dominican Republic. Association
  for Computational Linguistics.

\bibitem[{Sanh et~al.(2022)Sanh, Webson, Raffel, Bach, Sutawika, Alyafeai,
  Chaffin, Stiegler, Scao, Raja et~al.}]{sanh2021multitask}
Victor Sanh, Albert Webson, Colin Raffel, Stephen~H Bach, Lintang Sutawika,
  Zaid Alyafeai, Antoine Chaffin, Arnaud Stiegler, Teven~Le Scao, Arun Raja,
  et~al. 2022.
\newblock Multitask prompted training enables zero-shot task generalization.
\newblock In \emph{ICLR}.

\bibitem[{Smith(2018)}]{smith2018disciplined}
Leslie~N Smith. 2018.
\newblock A disciplined approach to neural network hyper-parameters: Part
  1--learning rate, batch size, momentum, and weight decay.
\newblock In \emph{CVPR}.

\bibitem[{Thorne et~al.(2018)Thorne, Vlachos, Christodoulopoulos, and
  Mittal}]{thorne2018fever}
James Thorne, Andreas Vlachos, Christos Christodoulopoulos, and Arpit Mittal.
  2018.
\newblock \href {https://doi.org/10.18653/v1/N18-1074} {{FEVER}: a large-scale
  dataset for fact extraction and {VER}ification}.
\newblock In \emph{Proceedings of the 2018 Conference of the North {A}merican
  Chapter of the Association for Computational Linguistics: Human Language
  Technologies, Volume 1 (Long Papers)}, pages 809--819, New Orleans,
  Louisiana. Association for Computational Linguistics.

\bibitem[{Wang et~al.(2021)Wang, Tang, Duan, Wei, Huang, Ji, Cao, Jiang, and
  Zhou}]{wang2020k}
Ruize Wang, Duyu Tang, Nan Duan, Zhongyu Wei, Xuanjing Huang, Jianshu Ji,
  Guihong Cao, Daxin Jiang, and Ming Zhou. 2021.
\newblock \href {https://doi.org/10.18653/v1/2021.findings-acl.121}
  {{K-Adapter}: {I}nfusing {K}nowledge into {P}re-{T}rained {M}odels with
  {A}dapters}.
\newblock In \emph{Findings of the Association for Computational Linguistics:
  ACL-IJCNLP 2021}, pages 1405--1418, Online. Association for Computational
  Linguistics.

\bibitem[{Wei et~al.(2022)Wei, Bosma, Zhao, Guu, Yu, Lester, Du, Dai, and
  Le}]{wei2021finetuned}
Jason Wei, Maarten Bosma, Vincent~Y Zhao, Kelvin Guu, Adams~Wei Yu, Brian
  Lester, Nan Du, Andrew~M Dai, and Quoc~V Le. 2022.
\newblock Finetuned language models are zero-shot learners.

\bibitem[{Yogatama et~al.(2014)Yogatama, Wang, Routledge, Smith, and
  Xing}]{yogatama2014dynamic}
Dani Yogatama, Chong Wang, Bryan~R. Routledge, Noah~A. Smith, and Eric~P. Xing.
  2014.
\newblock \href {https://doi.org/10.1162/tacl_a_00175} {Dynamic language models
  for streaming text}.
\newblock \emph{Transactions of the Association for Computational Linguistics},
  2:181--192.

\bibitem[{Zhang and Choi(2021)}]{zhang2021situatedqa}
Michael Zhang and Eunsol Choi. 2021.
\newblock \href {https://doi.org/10.18653/v1/2021.emnlp-main.586}
  {{S}ituated{QA}: Incorporating extra-linguistic contexts into {QA}}.
\newblock In \emph{Proceedings of the 2021 Conference on Empirical Methods in
  Natural Language Processing}, pages 7371--7387, Online and Punta Cana,
  Dominican Republic. Association for Computational Linguistics.

\end{thebibliography}
\bibliographystyle{acl_natbib}
\clearpage
\appendix


\section{Examples of \textsc{TWiki-Diffsets} and \textsc{TWiki-Probes}}
Figure \ref{fig:twiki_diffsets} shows the examples of \textsc{TWiki-Probes} which is either an 
updated or a new piece of information. By comparing consecutive snapshots of Wikipedia corpus, we keep track of changed information. Table \ref{table:twiki_probes_example} shows the examples of \textsc{Changed} factual instances \textsc{TWiki-Probes}, which is aligned with the corresponding sentence in \textsc{TWiki-Diffsets}.
\begin{figure}[t!]
    \centering
    \begin{subfigure}[b]{0.5\textwidth}
    \includegraphics[width=\textwidth]{figures/Figure2_1.pdf}
    \caption{\textsc{Information Update}}
    \end{subfigure}
    \begin{subfigure}[b]{0.5\textwidth}
    \includegraphics[width=\textwidth]{figures/Figure2_2.pdf}
    \caption{\textsc{New Information}}
    \end{subfigure}
    \caption{Examples of \textsc{TWiki-Diffsets} constructed from comparing November 2021 and December 2021 Wikipedia Dumps. (a) shows an instance of information update and (b) shows an instance of new information.}
    \label{fig:twiki_diffsets}
\vspace{-5mm}
\end{figure}

\begin{table*}[ht!]
    \centering
    \fontsize{10}{12}\selectfont
    \caption{Examples of successful alignment between \textsc{Changed} factual instances from \textsc{TWiki-Probes-0910} and articles from \textsc{TWiki-Diffsets-0910}. The alignment is considered successful because for the given factual instance, the \texttt{Subject} matches the title of the Wikipedia page and the \texttt{Object} exists in the article.}
    \resizebox{1\textwidth}{!}{\begin{tabular}{llll}
    \toprule
        \textbf{Subject}  & \textbf{Relation} & \textbf{Object} &{\textbf{Corresponding Sentence in Wikipedia}}\\
    \midrule
    Carlo Alighiero     & place of death   & Rome  & [...] \textbf{Carlo Alighiero} died in \textbf{Rome} on 11 September 2021 at the age of 94.[...]   \\
    \midrule
    Shang-Chi and the & instance of & Film & [...] \textbf{Shang-Chi and the Legend of the Ten Rings} is a 2021 American \\
    Legend of the Ten Rings & & & superhero \textbf{film} based on Marvel Comics featuring the character Shang-Chi.[...] \\
    \midrule
    Out of Shadows & language of work or name & Spanish & [...] \textbf{It} was later translated into Portuguese, Turkish and \textbf{Spanish.}[...]  \\
    \midrule
    Mario Chalmers & member of sports team & Indios & [...] On September 27, 2021, \textbf{Chalmers} signed with \textbf{Indios de Mayagüez} \\
     & & de Mayaguez & of the Baloncesto Superior Nacional.[...] \\
    \bottomrule
    \end{tabular}}
    \vspace{-2mm}
    \label{table:twiki_probes_example}
\end{table*}

\section{Details of Entity Types of \texttt{Subject} and \texttt{Relation}}
\label{appen:entity_type}
\begin{figure}[ht!]
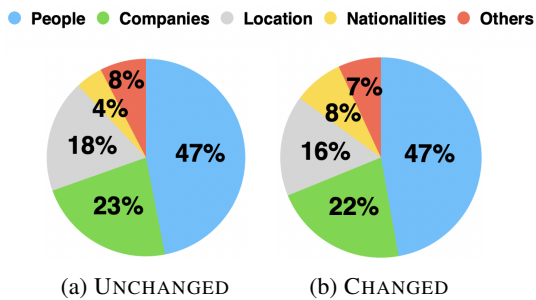

    \centering
    \begin{subfigure}[b]{0.92\columnwidth}
    \includegraphics[width=\textwidth]{figures/object_3.pdf}
    \end{subfigure}
    \hspace{5em}
    \begin{subfigure}[b]{0.35\columnwidth}
    \includegraphics[width=\textwidth]{figures/object_1.pdf}
    \caption{\textsc{Unchanged}}
    \end{subfigure}
    \hspace{0.5em}
    \begin{subfigure}[b]{0.35\columnwidth}
    \includegraphics[width=\textwidth]{figures/object_2.pdf}
    \caption{\textsc{Changed}}
    \end{subfigure}
    \hspace{0.5em}
\caption{Entity types of \texttt{Subject} and \texttt{Object} in \textsc{TWiki-Probes}.}
\label{fig:entity_stat}
\vspace{-5mm}
\end{figure}

Figure \ref{fig:entity_stat} shows the ratio of different entity types of \texttt{Subject} and \texttt{Object} of \textsc{Unchanged} and \textsc{Changed}.

\section{Details of \texttt{Relation} Distribution}
\begin{figure*}[ht!]
    \centering
    \begin{subfigure}[b]{0.82\textwidth}
    \includegraphics[width=\textwidth]{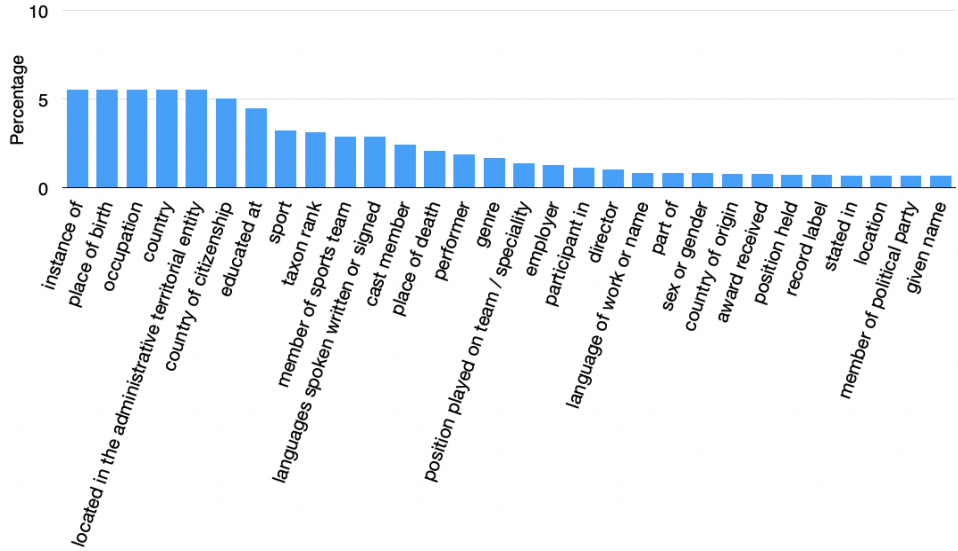}
    \caption{\textsc{Unchanged}}
    \end{subfigure}
    \begin{subfigure}[b]{0.82\textwidth}
    \includegraphics[width=\textwidth]{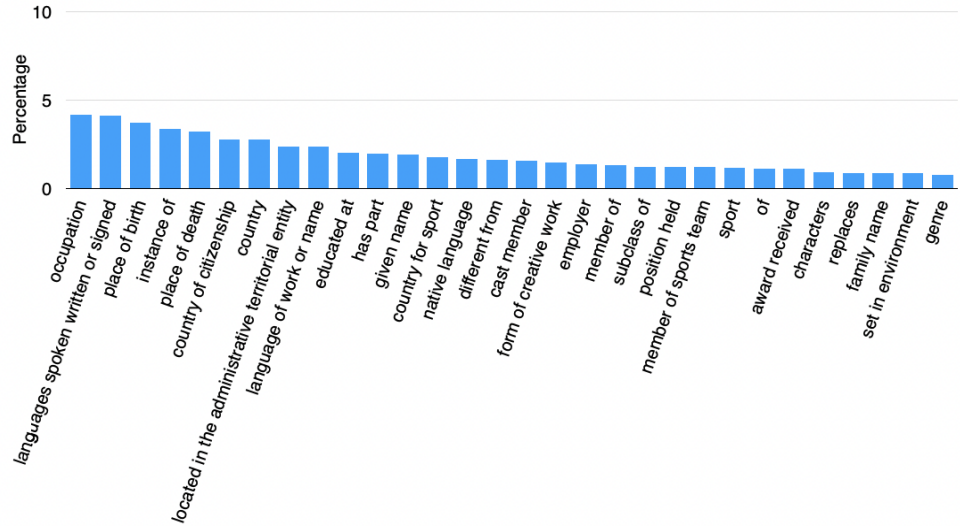}
    \caption{\textsc{Changed}}
    \end{subfigure}
\caption{\textsc{TWiki-Probes} distribution of the top 30 \texttt{Relation}.}
\label{fig:relation_stat_appendix}
\end{figure*}

The distribution of \texttt{Relation} for \textsc{Unchanged}, \textsc{Changed} factual instances in \textsc{TWiki-Probes} is shown in Figure \ref{fig:relation_stat_appendix}.
\label{appen:relation}

\section{Continual Pretraining and Light Tuning Configuration}
\label{appen:config}
For each LM update, we use 8 32GB V100 GPUs with a global batch size of 64 and a fixed input sequence length of 512. We use the max learning rate of 1e-4 and one cycle learning rate scheduling policy~\citep{smith2018disciplined}.
For light-tuning, the training is done for only one epoch with a learning rate of 1e-5 and a batch size of 32. Input and output sequence lengths are set to 25. For continual learning-based methods, we unfreeze all of the parameters during light-tuning, following \citet{jang2021towards}. 

\section{Light-tuning results with \textsc{TWiki-Probes}}
\label{appen:f1}

\begin{table*}[ht!]
\centering
\caption{Light-tuning perplexity of LMs measured on \textsc{TWiki-Probes}.}
    \resizebox{0.9\textwidth}{!}{\begin{tabular}{ccccccccc}
    \toprule
    \multicolumn{1}{l}{} & \multicolumn{2}{c}{TWiki-Probes-0809} & \multicolumn{2}{c}{TWiki-Probes-0910} & \multicolumn{2}{c}{TWiki-Probes-1011} &
    \multicolumn{2}{c}{TWiki-Probes-1112}
    \\ \cmidrule(lr){2-3} \cmidrule(lr){4-5} \cmidrule(lr){6-7} \cmidrule(lr){8-9}  & \textbf{Un} & \textbf{C} & \textbf{Un} & \textbf{C} & \textbf{Un} & \textbf{C} & \textbf{Un} & \textbf{C} \\
    \midrule
    \textsc{Initial} & 116.99 & 142.58 & \textbf{108.89} & 167.82 & 106.14 & 172.18 & \textbf{114.64} & 177.02  \\ \midrule
    \textsc{Full} & 124.37 & 145.89 & 112.51 & 172.70 & \textbf{105.09} & 164.59 & 118.54 & 164.17    \\
    \textsc{Diff} &120.52 & \textbf{116.44} & 125.80 & \textbf{142.82} & 132.83 & 156.60 & 144.61 & 164.34 \\ \midrule
    \textsc{RecAdam} & 122.58 & 118.14 & 125.90 & 143.65 & 137.15 & \textbf{148.24} & 144.76 & 159.52 \\
     \textsc{Mix-review} & 116.53 & 121.57 & 119.39 & 154.72 & 119.16 & 157.59 & 118.64 & \textbf{145.29} \\
     \textsc{LoRA} & 123.62 & 130.41 & 115.54 & 156.07 & 115.26 & 165.51 & 122.11 & 169.59 \\
    \textsc{K-Adapter} & \textbf{115.93} & 134.46 & 116.27 & 154.11 & 110.17 & 158.21 & 117.22 & 167.44 \\
    \bottomrule
    \end{tabular}}
\label{table:gpt2_light-tune}
\vspace{-5mm}
\end{table*}

\begin{table*}[ht!]
\centering
\caption{F1 score result of LMs on \textsc{TWiki-Probes} after light-tuning.}
    \resizebox{0.9\textwidth}{!}{\begin{tabular}{ccccccccc}
    \toprule
    \multicolumn{1}{l}{} & \multicolumn{2}{c}{TWiki-Probes-0809} & \multicolumn{2}{c}{TWiki-Probes-0910} & \multicolumn{2}{c}{TWiki-Probes-1011} &
    \multicolumn{2}{c}{TWiki-Probes-1112}
    \\ \cmidrule(lr){2-3} \cmidrule(lr){4-5} \cmidrule(lr){6-7} \cmidrule(lr){8-9}  & \textbf{Un} & \textbf{C} &  \textbf{Un} & \textbf{C} & \textbf{Un} & \textbf{C} & \textbf{Un} & \textbf{C} \\
    \midrule
    \textsc{Initial} & 6.98 & 3.19 & 7.26 & 3.35 & 7.27 & 2.74 & 6.84 & 2.82 \\ \midrule
    \textsc{Full} & 4.68 & 2.45 & 5.62 & 3.06 & 7.12 & 2.25 & 4.28 & 2.30 \\
    \textsc{Diff} & 7.51 & 4.38 & 6.91 & \textbf{4.46} & 5.24 & 2.65 & 5.45 & \textbf{4.38} \\ \midrule
    \textsc{RecAdam} & 5.74 & 3.79 & 6.31 & 3.86 & 4.47 & 2.43 & 5.09 & 3.68 \\
    \textsc{Mix-review} & 7.12 & 3.31 & 6.16 & 3.56 & 6.63 & 2.08 & 6.84 & 3.67\\
    \textsc{LoRA} & 7.36 & \textbf{4.48} & 7.23 & 3.89 & 7.19 & 3.87 & 6.82 & 3.81 \\
    \textsc{K-Adapter} & \textbf{7.54} & 3.99 & \textbf{7.34} & 3.73 & \textbf{7.38} & \textbf{3.91} & \textbf{6.87} & 3.30  \\
    \bottomrule
    \end{tabular}}
\label{table:gpt2_f1}
\vspace{-5mm}
\end{table*}

Using the pre-defined templates of LAMA~\citep{petroni2019language} seems to be an option, but we find that those templates do not fit well to our experiments because there is a considerable distribution gap between LAMA and \textsc{TWiki-Probes}; over half of the instances of \textsc{TWiki-Probes} are filtered out to apply the templates, especially for \textsc{Changed}.

Instead, to alleviate the distributional shift that causes high zero-shot perplexity, we \textit{light-tune} the LMs on 500 instances randomly sampled from WikiData that do not overlap with instances from \textsc{TWiki-Probes} (details in Appendix \ref{appen:lighttune}). Unlike finetuning, \textit{light-tuning} lets the LM only learn the input and output distribution of the task, avoiding the problem of test-train overlap pointed out by \citet{lewis-etal-2021-question}. Table \ref{table:gpt2_light-tune} shows the results of light-tuning, which demonstrate a similar trend as the zero-shot performance. 
Although light-tuning avoids the problem of test-train overlap, results are largely affected by the sampled instances for tuning, so a zero-shot evaluation setting is preferred for reliability. 

Many knowledge-intensive tasks such as closed-book question answering~\citep{roberts2020much, petroni2020kilt, jang2021towards} or slot filling \cite{petroni2020kilt} use accuracy, EM, or F1 score to evaluate the task. We also show the F1 score on \textsc{TWiki-Probes} in Table \ref{table:gpt2_f1}. Overall trend is consistent with zero-shot perplexity metric; \textsc{K-Adapter} shows robust performance for both \textsc{Unchanged} and \textsc{Changed}.

\section{Light-Tuning Data}
\label{appen:lighttune}

\begin{table}[ht!]
\centering
    \caption{Statistics of the data used for Light-Tuning}
\resizebox{\columnwidth}{!}{\begin{tabular}{ccccc}
\toprule
& \textbf{Size} & \begin{tabular}[c]{@{}c@{}}\textbf{\# of} \\ \textbf{Relation} \end{tabular} & \begin{tabular}[c]{@{}c@{}c@{}}\textbf{Maximum}\\ \textbf{Repetition}\\ \textbf{of Relation}\end{tabular} & \begin{tabular}[c]{@{}c@{}}\textbf{\# of} \\ \textbf{Subject} \end{tabular}  \\
\midrule
\textsc{Unchanged} & 500  & 102  & 58 & 499           \\
\textsc{Changed}   & 500  & 140  & 31 & 500           \\
\bottomrule
\end{tabular}}
    \label{table:light-tuning datasets}
\end{table}

We sample 500 instances from WikiData for each time step that do not overlap with instances from \textsc{TWiki-Probes} for each factual instance category. During sampling, we keep the distribution of each \texttt{Relation} proportional to the original distribution. Table \ref{table:light-tuning datasets} shows the size and distribution of \texttt{Relation} of light-tuning datasets.

\end{document}